\documentclass[11pt]{article}

\usepackage[preprint]{acl}

\usepackage{times}
\usepackage{latexsym}

\usepackage[T1]{fontenc}

\usepackage[utf8]{inputenc}

\usepackage{microtype}

\usepackage{inconsolata}

\usepackage{graphicx}

\usepackage{amsmath}
\usepackage{amssymb}
\usepackage{booktabs}
\usepackage{multirow}
\usepackage{enumitem}
\usepackage{tcolorbox}

\title{Output Vector Editing for Memorization Mitigation in Large Language Models}

\author{
Ahmad Dawar Hakimi\textsuperscript{1,2,3,4,*} \quad
Kaiwei Lei\textsuperscript{1,2,*} \quad
Isabelle Augenstein\textsuperscript{3,5} \quad
Hinrich Schütze\textsuperscript{1,4}
\\[2pt]
\textsuperscript{1}Center for Information and Language Processing, LMU Munich, Germany \\
\textsuperscript{2}Department of Computer Science, University of Copenhagen, Denmark \\
\textsuperscript{3}Munich Center for Machine Learning, Germany \\
\textsuperscript{4}Pioneer Centre for AI, Denmark \\
\textsuperscript{5}BMW Group, Germany \\
\texttt{adhakimi@cis.lmu.de} \\
\textsuperscript{*}Equal contribution.
}

\newcounter{notecounter}
\newcommand{\enotesoff}{\long\gdef\enote##1##2{}}

\enotesoff

\begin{document}
\maketitle
\begin{abstract}
Large language models can reproduce training text, creating
privacy, copyright, and security risks. We propose Output
Vector Editing (OVE), a post-hoc method for suppressing text
continuations that a model reproduces from a given
prefix. Under superposition, a neuron can contribute to
multiple features. Yet some neuron-level mitigation methods
suppress localized neurons by zeroing their activations,
removing each neuron's entire contribution. OVE instead
edits their output vectors, combining (i) two-stage
localization of MLP layers and neurons
with (ii) full-vocabulary least-squares updates. Its multi-anchor variant, OVE-MA, adaptively selects continuation positions expected to provide stronger suppression relative to the required weight change. We evaluate recovery and locality across four models (360M--7B parameters), using autoregressively validated continuations and recovery tests with greedy decoding, beam search, and sampling. On 634 OLMo-7B pairs, OVE-MA with at most five output-vector updates per request leaves 24.45\% recoverable in these tests, compared with 32.02\% for the closest prior post-hoc editing method, REVS, while using approximately 75 times fewer update operations. These results show that adaptive position selection can concentrate effective suppression into a small number of weight updates. OVE edits weights without gradients and requires no additional intervention during inference.
\end{abstract}

\section{Introduction}
\label{sec:intro}

\begin{figure}[t]
    \centering
    \includegraphics[width=\linewidth]{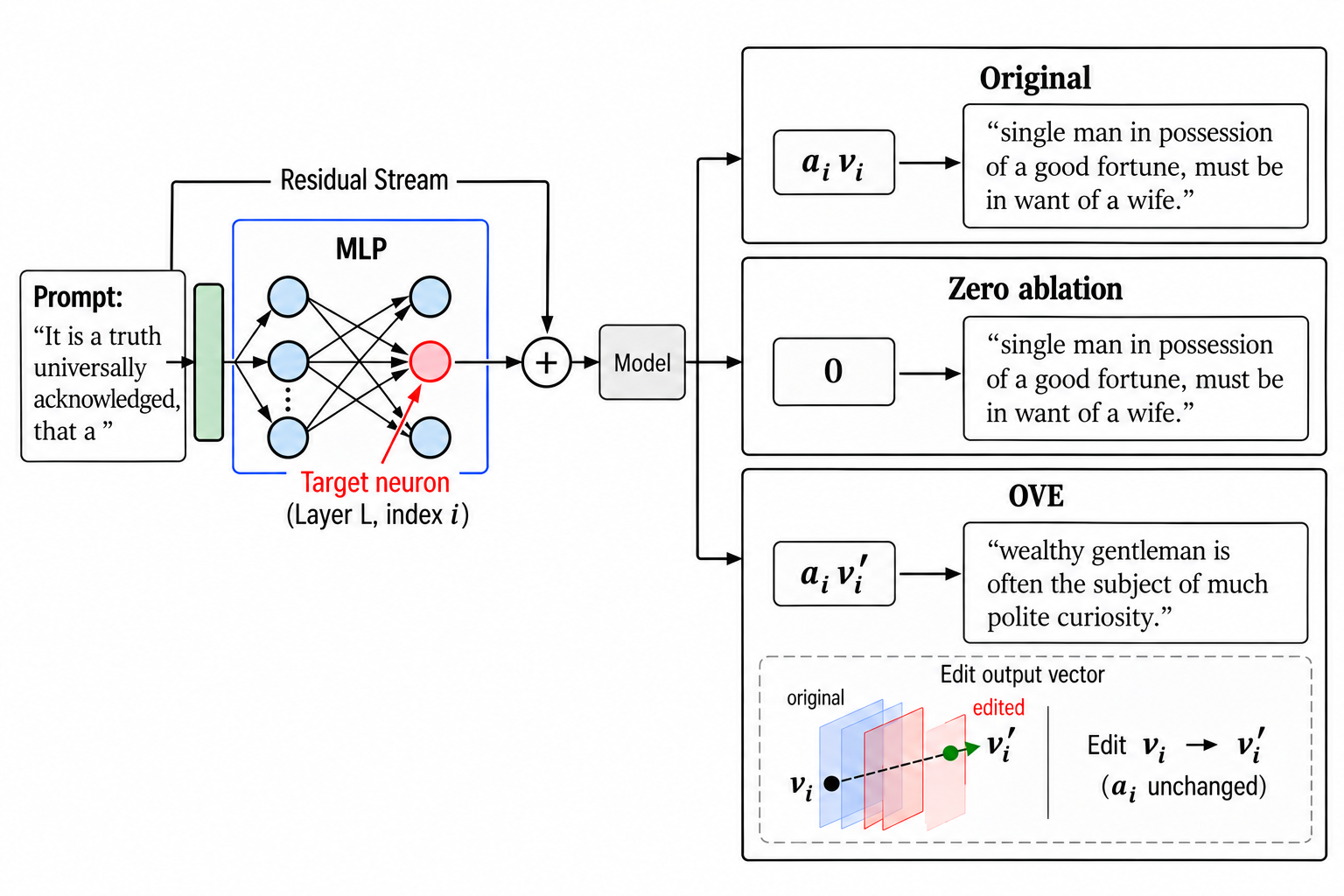}
    \caption{Comparison of interventions on a target MLP neuron for the prompt ``It is a truth universally acknowledged, that a''. The neuron originally contributes $a_i\mathbf{v}_i$ to the residual stream. \textbf{Original:} the model reproduces the target continuation. \textbf{Zero ablation:} removing the neuron's contribution still yields the same continuation in this example. \textbf{OVE:} we edit the output vector $\mathbf{v}_i \rightarrow \mathbf{v}'_i$ while keeping the activation $a_i$ unchanged, replacing the contribution with $a_i\mathbf{v}'_i$ and producing a different continuation.}
    \label{fig:overview}
\end{figure}

Large language models can reproduce sequences from their training data \citep{carlini2021extracting,carlini2023quantifying}, creating risks when the content is private, copyrighted, or otherwise sensitive \citep{hartmann2023sok,schwinn2023adversarial}. We study a post-hoc white-box setting in which a model has been observed to reproduce a text continuation from a given prefix, and the goal is to suppress that reproduction. We refer to such outputs as \emph{reproduced continuations}. Existing parameter-editing methods are largely designed for factual associations, where the content to edit and its desired replacement are specified in advance \citep{meng2022locating,meng2023massedit}. Reproduced textual continuations pose a different problem: their prefixes are free form, the relevant prediction position must be identified within the sequence, and there is no predefined replacement. Existing post-hoc approaches for this setting either remove localized model components or edit their vocabulary-space contributions \citep{wu2023depn,sakarvadia2025mitigating,ruzzetti2025private, ashuach2025revs}.

We propose \textbf{Output Vector Editing (OVE)}
(Fig.~\ref{fig:overview}), a post-hoc method that localizes MLP neurons and modifies their output vectors. Its two-stage localization selects MLP layers by their target-token projections and neurons by their contributions to the residual stream. Each update solves a full-vocabulary least-squares problem that requests a change to one token projection while penalizing changes to the others. The operator supports suppression of the target token or redirection toward an alternative token. The edits are stored in the model weights and require no additional intervention during inference.

The effect of an edit also depends on the targeted continuation position. We derive the fraction of a requested projection change achieved for each token and use it to score positions by predicted suppression relative to update size. OVE's multi-anchor configuration, \textbf{OVE-MA}, iteratively selects a position, applies updates, and rescores the remaining positions on the edited model, within a fixed budget of output-vector updates.

Overall, we contribute a targeted output-vector editing method, an adaptive strategy for selecting continuation positions, and an evaluation of their effectiveness and locality across models and decoding procedures. We first mine candidate continuations using teacher-forced next-token predictions, then verify that the model actually reproduces them when generating from the original prefix. On OLMo-7B, only 634 of 6,831 mined candidates pass this autoregressive validation. On these validated pairs, OVE-MA achieves lower recovery than REVS with substantially fewer output-vector updates, showing that selecting where to intervene can make suppression considerably more
effective.
Across models and longer edit sequences, the balance between suppression and locality varies, highlighting the importance of evaluating both.


\section{Related Work}
\label{sec:related}

\textbf{Localization and mechanisms of memorization.}
Transformer MLPs have been interpreted as key-value memories whose output vectors contribute to vocabulary predictions \citep{geva2021transformer,geva2022transformer}. Localization studies examine parameter subsets, neurons, and circuits associated with memorization \citep{maini2023can,chang2024localization,lasy2025understanding}. Localized neurons need not be specific to one sequence and can encode multiple features \citep{chang2024localization,cunningham2023sparse}. Memorized generation can also depend on distributed states and general language modeling capabilities \citep{huang2024demystifying}. These findings motivate modifying what a localized neuron writes to the residual stream rather than removing its entire contribution. OVE therefore edits the neuron's output vector rather than zeroing its activation.

\paragraph{Parameter editing, unlearning, and targeted memorization mitigation.} Our setting is closely related to language-model unlearning, particularly work on memorized textual sequences \citep{sakarvadia2025mitigating,barbulescu2024each,zhang2025getconfused}. OVE adopts a narrower objective: given a reproduced prefix--continuation pair $(x,y)$, it suppresses $y$ from $x$ without claiming removal of the underlying information. Unlike factual editing methods such as ROME and MEMIT \citep{meng2022locating,meng2023massedit}, OVE targets free-form continuations without a predefined replacement fact. DEPN removes localized neuron contributions \citep{wu2023depn}, while PISCES removes concept-associated parameter directions \citep{gurarieh2025precise}; OVE instead modifies localized output vectors. Private Memorization Editing similarly targets known memorized sensitive content \citep{ruzzetti2025private}. Among prior methods, REVS is closest: like OVE, it edits MLP output vectors in vocabulary space \citep{ashuach2025revs}. The methods differ in their localization and editing procedures: REVS iteratively moves token projections into specified rank ranges, whereas OVE directly specifies a target-token projection and computes a full-vocabulary least-squares update. OVE-MA additionally selects continuation positions adaptively.


\paragraph{Preventing and mitigating memorized generation.}
Training-time approaches reduce memorization through deduplication \citep{lee2022deduplicating,kandpal2022deduplicating} or modified training objectives \citep{hans2024goldfish}. ParaPO uses post-training preference optimization to favor paraphrases over verbatim text \citep{chen2025parapo}. At inference time, MemFree blocks training-set $n$-grams \citep{ippolito2023preventing}, TokenSwap selectively replaces function-word probabilities using a smaller auxiliary model \citep{prashant2025tokenswap}, and CP-Fuse adaptively combines model logits \citep{abad2024cpfuse}. Anchored Decoding constrains generation relative to a reference model trained on permissively licensed data \citep{he2026anchoreddecoding}, while activation steering modifies hidden states during generation \citep{suri2025mitigating}. OVE instead modifies model weights post hoc: once an identified continuation is edited, subsequent generation requires no decoding constraint, auxiliary model, or activation intervention.

\section{Method}
\label{sec:method}

We consider a prefix--continuation pair $(x,y)$ for which the model reproduces $y$ from $x$. We refer to $y$ as the target continuation. Our goal is to suppress this reproduction under the tested decoding procedures. OVE uses a selected continuation position to localize MLP neurons and edit their output vectors. Its multi-anchor configuration, OVE-MA, repeats this procedure at adaptively selected positions under a fixed update budget. Each modification of one neuron's output vector counts as one update.


\subsection{Why edit output vectors, not activations?}
\label{sec:motivation}

An MLP layer's output is a sum of neuron contributions
\citep{geva2021transformer,cunningham2023sparse}:
\begin{equation}
\operatorname{MLP}(\mathbf{h}) =
\sum_{i=1}^{d_{\mathrm{mlp}}} a_i(\mathbf{h})\mathbf{v}_i ,
\label{eq:decomposition}
\end{equation} 
where $\mathbf{h}\in\mathbb{R}^d$ is the layer input, $d_{\mathrm{mlp}}$ is the number of neurons, $a_i(\mathbf{h})$ is neuron $i$'s scalar activation, and $\mathbf{v}_i\in\mathbb{R}^d$ is its output vector, stored as row $i$ of $\mathbf{W}_{\mathrm{out}}$. For gated MLPs, $a_i(\mathbf{h})$ includes the gate and can be negative. We omit the argument $\mathbf{h}$ below when the input is clear from context. The activation scales the neuron's contribution; the output vector determines what it writes to the residual stream.

Prior work uses neuron ablation to study memorization \citep{maini2023can,chang2024localization}. Setting $a_i=0$ removes the neuron's entire contribution $a_i\mathbf{v}_i$. Under superposition, a neuron can contribute to multiple features \citep{cunningham2023sparse,gurnee2023finding}, so ablation removes its contribution to all of them.

OVE instead replaces $\mathbf{v}_i$ with $\mathbf{v}'_i$. For a fixed layer input, $a_i$ remains unchanged and the contribution becomes $a_i\mathbf{v}'_i$. The weight edit applies wherever the neuron is used. Other vocabulary projections and downstream activations can still change, so we evaluate locality empirically.

 
\subsection{Candidate mining and validation}
\label{sec:mining}

We use a two-stage procedure to identify prefix--continuation pairs that the unedited model demonstrably reproduces. First, we efficiently screen corpus text under teacher forcing. For each sentence, teacher forcing yields a next-token prediction at every position, conditioned on the preceding ground-truth tokens. We compare these predictions with the reference tokens in every contiguous 20-token window, compute BLEU for each window, and retain the window with the highest BLEU if it exceeds 0.5. The preceding text forms the prefix $\mathbf{x}$ and the selected window the candidate continuation $\mathbf{y}$.
Teacher forcing can overestimate reproduction because every prediction is conditioned on the preceding ground-truth tokens. We therefore validate each candidate by generating 20 tokens greedily from $\mathbf{x}$ alone. We call the BLEU between this generation and $\mathbf{y}$ autoregressive BLEU (AR-BLEU) and retain pairs with AR-BLEU $\geq 0.75$ as our operational criterion for approximate memorization. Only these validated pairs are used for editing and recovery evaluation.

Where the model's training corpus is available, we mine from it. For Llama2-7B~\cite{touvron2023llama2}, whose 
training corpus is unavailable, we instead use SlimPajama~\cite{shen2023slimpajama}, as a proxy corpus; these examples therefore establish reproduction but not training-set membership. BLEU is computed on token IDs, and generated continuations are scored only before the first end-of-sequence token. App.~\ref{app:mining} and~\ref{app:cross_model_models} provide details.

\subsection{Anchors and two-stage localization}
\label{sec:localization}

\paragraph{Anchors.}

An anchor $j$ is a token position within the target continuation $y$. At anchor $j$, the model is given the original prefix $\mathbf{x}$ followed by the preceding continuation tokens $\mathbf{y}_{<j}$, and the target is the next reference token $t=y_j$. Positions are eligible unless their target belongs to a fixed token-ID set constructed from stopwords, punctuation, and single letters (App.~\ref{app:stopword_filter}). The \emph{memorization anchor} $j_{\mathrm{mem}}$ is the first eligible position. Single-anchor OVE edits at this position, whereas OVE-MA selects among all eligible positions.

All localization quantities are evaluated at the final token of this anchor prefix. Let $\mathbf{W}_U=[\mathbf{u}_1,\ldots,\mathbf{u}_{|V|}] \in\mathbb{R}^{d\times |V|}$ be the vocabulary-mean-centered unembedding matrix, where $V$ is the vocabulary. We project directly through $\mathbf{W}_U$, omitting the model's final normalization and unembedding bias. These projections provide intermediate vocabulary scores.

\paragraph{Step 1: Select MLP layers.}
For the MLP output $\mathbf{m}^{\ell}$ at layer $\ell$, we define
\begin{equation}
\operatorname{LI}_{\ell}(t)=
\frac{\mathbf{u}_t^\top\mathbf{m}^{\ell}}
{\max_{w\in V}\mathbf{u}_w^\top\mathbf{m}^{\ell}} .
\label{eq:layer_importance}
\end{equation}
This measures the target-token projection relative to the layer's largest vocabulary projection. We retain up to five MLP layers with the highest scores above 0.2.



\paragraph{Step 2: Select neurons.}
Within each selected layer, we score neurons by the proximity of their contributions to the residual stream after the MLP block \citep{ferrando2024information}:
\begin{equation}
\begin{aligned}
D_i &=
\left\|a_i\mathbf{v}_i
-\mathbf{r}_{\mathrm{post}}^\ell
+\epsilon_{\mathrm{dist}}\mathbf{1}\right\|_1,\\
s_i &=
\max\!\left(
\epsilon_{\mathrm{score}},
\|\mathbf{r}_{\mathrm{post}}^\ell\|_1-D_i
\right),
\end{aligned}
\label{eq:neuron_score}
\end{equation}
where $\mathbf{r}_{\mathrm{post}}^\ell$ is the post-MLP residual and $\mathbf{1}$ is the all-ones vector. Contributions closer to this residual receive higher scores, subject to the score floor. Numerical constants and score normalization are specified in App.~\ref{app:baseline_implementation}.

We take the top $k=5$ neurons per selected layer, giving at most 25 candidates. To check their association with the target token, we retain a candidate  neuron $i$ only if $t$ is among the 50 highest or 50 lowest coordinates of $\mathbf{W}_U^\top\mathbf{v}_i$. This filter uses the output vector without multiplying by its activation. OVE-MA's update budget limits how many retained vectors are edited. If no candidates remain, no edit is available at that anchor. Coverage is the fraction of pairs receiving at least one output-vector update.

\subsection{Output vector editing}
\label{sec:editing}

Let $\mathbf{u}_s=\mathbf{W}_U[:,s]\in\mathbb{R}^d$ denote the unembedding column for edit token $s$. For suppression, $s$ is the target token $t$; for redirection, $s$ is a substitute token. The vector
\[
\mathbf{W}_U^\top\mathbf{v}_i\in\mathbb{R}^{|V|}
\]
contains neuron $i$'s direct vocabulary projections. For each localized neuron $i$, OVE specifies a desired value $q_i$ for the $s$-th projection $\mathbf{u}_s^\top\mathbf{v}_i$. This corresponds to the requested change
\[
\delta_i=q_i-\mathbf{u}_s^\top\mathbf{v}_i
\]
in coordinate $s$, while ideally leaving the other vocabulary coordinates unchanged. \S\ref{sec:modes} defines $q_i$ for suppression; App.~\ref{app:modes} defines it for the redirection modes.

Let $\mathbf{e}_s\in\mathbb{R}^{|V|}$ denote the vocabulary basis vector for token $s$. The desired vocabulary-space change is therefore $\delta_i\mathbf{e}_s$. OVE finds the output-vector update whose vocabulary projection best approximates this change. For full-row-rank
$\mathbf{W}_U$, the solution is
\begin{equation}
\begin{aligned}
\Delta\mathbf{v}_i
&=\arg\min_{\mathbf{z}}
\left\|\mathbf{W}_U^\top\mathbf{z}
-\delta_i\mathbf{e}_s\right\|_2^2\\
&=\delta_i
(\mathbf{W}_U\mathbf{W}_U^\top)^{-1}\mathbf{u}_s .
\end{aligned}
\label{eq:objective}
\end{equation}
We store $\mathbf{v}'_i=\mathbf{v}_i+\Delta\mathbf{v}_i$ in the model weights. Each nonzero vector update changes one row of $\mathbf{W}_{\mathrm{out}}$ and has rank one. The edits require no gradients or additional intervention during inference.

\paragraph{Requested and realized changes.}
The least-squares update need not attain $q_i$ exactly. In exact arithmetic,
\begin{equation}
\begin{aligned}
\mathbf{u}_s^\top\Delta\mathbf{v}_i &= c_s\delta_i,\\
c_s &=
\mathbf{u}_s^\top
(\mathbf{W}_U\mathbf{W}_U^\top)^{-1}\mathbf{u}_s .
\end{aligned}
\label{eq:realization}
\end{equation}
The fraction $c_s\in[0,1]$ measures how much of the requested change is realized. It depends only on the edit token and the unembedding matrix. For suppression, $s=t$, giving the target-specific fraction $c_t$. Other vocabulary projections can also change. These expressions describe direct projections; effects on final predictions additionally depend on activations, normalization, and subsequent computation. See App. \ref{app:operator_diagnostics} for further operator properties.

\subsection{Suppression objective}
\label{sec:modes}

Our main configuration targets the target token, setting $s=t$ and requesting
\begin{equation}
q_i =
-\beta\,\operatorname{sign}(a_i)\,
\left|\mathbf{u}_t^\top\mathbf{v}_i\right|,
\label{eq:suppress}
\end{equation}
where $a_i$ is the neuron's activation at the last position of the anchor prefix. The parameter $\beta$ controls the request strength; we use $\beta=20$.

Accounting for the activation sign makes the requested contribution $a_iq_i$ nonpositive, including when $a_i$ is negative. However, the realized contribution can remain positive when $c_t$ is small. App.~\ref{app:operator_diagnostics} gives the condition under which its sign changes.

We also evaluate EOS, Redact, and Next-best redirection modes as ablations. Their definitions and results are provided in App.~\ref{app:modes}.

\subsection{Multi-anchor editing}
\label{sec:multi_anchor}

OVE-MA selects anchors by predicted suppression relative to update size and rescores the remaining anchors after editing.

\paragraph{Anchor score.}
For each eligible anchor $j$, we localize neurons on the current model using the original prefix $\mathbf{x}$ followed by the preceding continuation tokens $\mathbf{y}_{<j}$, with target $t=y_j$. Let $\mathcal{N}_j$ denote the localized neuron set and $\Delta\mathbf{v}_i$ the proposed suppression update for neuron $i$ (\S\ref{sec:editing}--\ref{sec:modes}). Stacking these updates as the rows of $\Delta\mathbf{W}_j$, we define
\begin{equation}
\begin{aligned}
R_j &=
-\sum_{i\in\mathcal{N}_j}
a_i\mathbf{u}_{y_j}^{\top}\Delta\mathbf{v}_i,\\
E_j &=
\frac{R_j}{\|\Delta\mathbf{W}_j\|_F}.
\end{aligned}
\label{eq:editability}
\end{equation}
Here, $R_j$ estimates the decrease in the selected neurons' direct target-token contribution, holding their activations fixed, while $E_j$ measures this predicted suppression relative to the size of the proposed weight update. We only consider anchors with a nonzero proposed update.

\paragraph{Selection and budget.}
The budget $B$ counts output-vector update
operations. Updating the same vector at a different anchor counts as a separate operation each time. Among unused anchors with localized neurons and $R_j>0$, we select the one with the highest $E_j$. We apply its updates in decreasing layer importance, then decreasing neuron score within each layer, up to the remaining budget. Each anchor's score is computed using all neurons in $\mathcal{N}_j$, even when the remaining budget allows updates to only a subset of them.

Each anchor is selected at most once. If budget remains, we re-localize and rescore all unused eligible anchors on the edited model. Editing stops when the budget is exhausted or no unused localizable anchor has $R_j>0$. Our main configuration uses $B=5$; we also evaluate $B=20$. App.~\ref{app:anchor_scoring} defines the score for redirection modes.

\section{Experimental Setup}
\label{sec:setup}
 
\subsection{Model and data}

We use OLMo-7B \citep{groeneveld2024olmo} as our primary benchmark because its weights and training corpus are publicly available, allowing us to identify reproduced continuations from training data and to run the full set of ablations and locality analyses. From 10M sentences across the Books and CC-En-Head partitions of Dolma~\citep{soldaini2024dolma}, we mine 6{,}831 candidates, of which 634 pass autoregressive validation (\S\ref{sec:mining}). We evaluate all 634 pairs, including 240 used during method development and examples examined in exploratory analyses. App.~\ref{app:mining} and~\ref{app:data_splits} describe mining and data selection.

To test whether the method transfers beyond this primary setting, we apply the same OVE configuration to 205 OLMo-1B pairs, 300 SmolLM-360M~\citep{allal2024SmolLM} pairs, and 300 Llama2-7B pairs. Because Llama2's training corpus is unavailable, we use SlimPajama as a proxy corpus; these experiments measure suppression of reproduced proxy-corpus continuations rather than confirmed training examples. App.~\ref{app:cross_model_models} provides details.

\subsection{Compared methods}
\label{sec:compared_methods}

Across all four models, we compare single-anchor OVE, OVE-MA with $B\in\{5,20\}$, REVS \citep{ashuach2025revs}, and zero ablation. OVE uses suppression unless stated otherwise, with OVE-MA at $B=5$ as our main configuration. Single-anchor OVE edits at the memorization anchor $j_{\mathrm{mem}}$, whereas OVE-MA adaptively selects among eligible continuation positions. We restore the original weights between independent editing requests. On OLMo-7B, we additionally evaluate anchor-selection and edit-mode ablations, random-neuron and activation-steering controls, and the first five and twenty updates of REVS' full update sequence. The latter are untuned truncation diagnostics, not optimized budgeted variants. Full definitions and implementation details are given in Apps.~\ref{app:anchor_scoring}, \ref{app:baseline_implementation}, and \ref{app:modes}.

\subsection{Recovery evaluation}
All recovery tests generate from the original prefix $\mathbf{x}$ and compare outputs with the reference continuation $\mathbf{y}$. \emph{Greedy residual} is the fraction of requests whose greedy output still has AR-BLEU $\geq 0.75$, while \emph{recovery} counts a request as recovered if any output in a common 21-output test spanning greedy decoding, beam search, and sampling reaches this threshold. Requests receiving no edit remain in the denominator and use the unedited model's outputs. We additionally report exact reproduction, edit coverage, and degeneration diagnostics. App.~\ref{app:recovery_protocol} gives the full decoding protocol.

\subsection{Locality and cumulative editing}
We evaluate side effects at three levels. First, \emph{same-target controls} test whether an edit disrupts other contexts requiring the same target tokens. Second, random contexts measure changes in the full next-token distribution using $\mathrm{KL}(p_{\mathrm{pre}}\|p_{\mathrm{post}})$. Third, we measure changes in WikiText-103~\citep{merity2017pointer} likelihood and performance on ~\citep{zellers2019hellaswag}, ARC-Challenge~\citep{clark2018arc}, and Wikidata QA~\citep{srivastava2023beyond}. All methods are compared on shared control and evaluation panels; Apps.~\ref{app:locality_protocol} and~\ref{app:performance_protocol} show construction and scoring details.

\paragraph{Cumulative editing.}
To test whether side effects accumulate, we apply the same 100 development requests to OLMo-7B in three seeded orders, without restoring the original weights. We compare OVE-MA with $B=5$ and $B=20$ against full REVS and an untuned truncation that applies only the first 20 updates of REVS' full update sequence. At $N\in\{1,10,25,50,100\}$, we reevaluate previously edited requests under greedy decoding and repeat the downstream-performance probes; at $N=100$, we additionally run the full 21-output recovery test. App.~\ref{app:cumulative} gives the full protocol.

\section{Results}
\label{sec:results}

We first examine how anchor selection and multi-anchor editing affect
suppression (\S\ref{sec:anchor_selection}--\ref{sec:multi-anchor-editing}),
then test recovery under additional decoding procedures
(\S\ref{sec:recovery}). We finally evaluate locality under independent
and cumulative edits (\S\ref{sec:tradeoff}) and ask whether the
resulting trade-offs transfer across models (\S\ref{sec:cross_model}).

\subsection{Anchor selection}
\label{sec:anchor_selection}

We compare three anchor policies under the same five-update budget: MEM edits at the first eligible position, RARE selects the eligible localizable position with the rarest target token, and EDIT selects the position with the highest predicted editability $E_j$. EDIT substantially improves suppression over MEM: recovery falls from 63.72\% to 50.00\% (Table~\ref{tab:anchor_selection}), a reduction of 13.72 percentage points (95\% paired bootstrap CI: $[8.83,18.61]$). This advantage persists when restricting the comparison to requests that both policies can edit.

RARE achieves full coverage but leaves more continuations recoverable than either MEM or EDIT, indicating that token rarity alone is not a useful proxy for editability. EDIT also causes more same-target damage than MEM (10.70\% versus 6.92\%), suggesting that positions at which the operator can suppress more effectively can also be less local. Anchor selection therefore exposes a suppression--locality trade-off rather than simply increasing the number of editable requests. App.~\ref{app:anchor_diagnostics} provides additional diagnostics.

\begin{table}[t]
\centering
\small
\setlength{\tabcolsep}{3pt}
\begin{tabular}{@{}lrrrr@{}}
\toprule
Policy &
Coverage $\uparrow$ &
\shortstack{Greedy\\residual $\downarrow$} &
Recovery $\downarrow$ &
\shortstack{Exact-20\\recovery $\downarrow$} \\
\midrule
MEM  & 92.3           & 41.48          & 63.72          & 31.07 \\
RARE & \textbf{100.0} & 65.30          & 80.76          & 46.06 \\
EDIT & \textbf{100.0} & \textbf{35.96} & \textbf{50.00} & \textbf{5.21} \\
\bottomrule
\end{tabular}
\caption{Anchor selection on all 634 OLMo-7B pairs with at most five updates per request. All values are percentages; bold marks the best value in each column. Recovery uses 21 outputs per pair, and unedited requests remain in the denominator.}
\label{tab:anchor_selection}
\end{table}

\subsection{Multi-anchor editing}
\label{sec:multi-anchor-editing}

We next test whether adapting the edit across multiple continuation positions improves suppression. The single-anchor OVE baseline here applies all retained updates at $j_{\mathrm{mem}}$ and is therefore not capped at five updates like MEM in Table~\ref{tab:anchor_selection}.

OVE-MA substantially outperforms single-anchor OVE on OLMo-7B: with $B=5$, recovery falls from 63.56\% to 24.45\% while coverage increases to all 634 requests (Table~\ref{tab:main}). 
OVE-MA also achieves lower recovery than full REVS on this model
(24.45\% versus 32.02\%; 95\% paired bootstrap CI for the
difference: [2.84, 12.15]),
while using about 5 rather than
376 output-vector updates per request, a roughly 75-fold reduction. Increasing the OVE-MA budget to $B=20$ further reduces recovery to 2.05\%.

\begin{figure}[t]
    \centering
    \includegraphics[width=\linewidth]{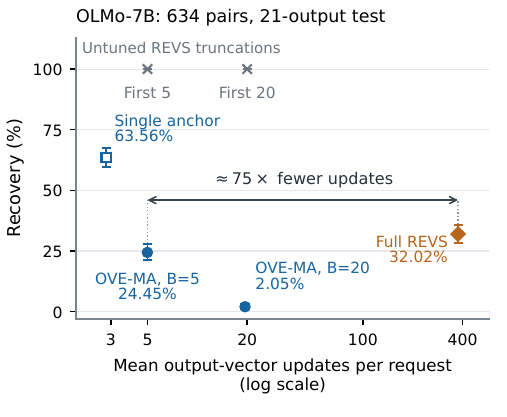}
    \caption{Recovery versus mean output-vector update operations on all 634 OLMo-7B pairs. The horizontal axis is logarithmic; points are evaluated configurations and bars show 95\% bootstrap intervals. REVS first-5 and first-20 are untuned truncations of its full update sequence. OVE-MA at $B=5$ uses approximately 75$\times$ fewer update operations than full REVS.}
    \label{fig:recovery_updates}
\vspace{-5mm}
\end{figure}

The methods also differ in where these updates occur: OVE variants concentrate 84--91\% of their update operations in the final eight OLMo-7B layers, compared with 48\% for REVS, whose updates are distributed more broadly across model depth (App.~\ref{app:layer_distribution}). Figure~\ref{fig:recovery_updates} summarizes the effectiveness--update trade-off.

\begin{table}[t]
\centering
\small
\setlength{\tabcolsep}{3pt}
\begin{tabular}{@{}lrrr@{}}
\toprule
Method &
\shortstack{Greedy\\residual $\downarrow$} &
Recovery $\downarrow$ &
\shortstack{Mean\\updates} \\
\midrule
OVE, single anchor & 41.01 & 63.56 & 2.81 \\
OVE-MA, $B=5$     & 14.51 & 24.45 & 4.99 \\
OVE-MA, $B=20$    & \textbf{1.10} & \textbf{2.05} & 19.41 \\
REVS, full        & 12.62 & 32.02 & 376.01 \\
\bottomrule
\end{tabular}
\caption{OLMo-7B results on all 634 pairs. Recovery metrics are percentages ($\downarrow$); bold marks the best suppression results among the primary weight-editing methods. Updates are mean output-vector operations per request.}
\label{tab:main}
\vspace{-5mm}
\end{table}

\paragraph{Mechanism controls.}
Zero ablation and random-neuron controls are substantially less effective than OVE's targeted updates. Activation steering reaches similar recovery to OVE-MA at $B=5$ but produces more repetitive degeneration. Full results are reported in App.~\ref{app:mechanism_controls}.

Applying only the first five or twenty updates from REVS' full
update sequence leaves all requests recoverable, showing that its existing update sequence cannot simply be truncated to OVE-sized budgets without retuning.

\paragraph{Edit modes.}
Alternative objectives can reduce recovery through early termination or repetitive generation rather than clean suppression; we therefore use direct suppression as the main configuration. The modes nevertheless exhibit complementary failure patterns (App.~\ref{app:modes}).

\subsection{Recovery under additional decoding}
\label{sec:recovery}

We next ask whether an edit that prevents greedy reproduction remains effective under alternative decoding procedures. Greedy decoding understates recoverability: for OVE-MA at $B=5$, recovery increases from 14.51\% under greedy decoding to 24.45\% under the full 21-output test, while REVS increases from 12.62\% to 32.02\%. REVS therefore appears more effective under greedy decoding but less effective once additional decoding strategies are considered. Expanding the test further on a fixed 100-request subset increases recovery for all methods, with the largest increase for REVS (App.~\ref{app:recovery_protocol}). These results show that conclusions about memorization mitigation depend on how aggressively recovery is tested. Failure under a finite set of decoding procedures does not establish that a continuation is unrecoverable.

\subsection{Locality and cumulative editing}
\label{sec:tradeoff}

OVE-MA's stronger suppression on OLMo-7B comes with greater target-specific side effects. On identical same-target controls, damage is 19.31\% for OVE-MA at $B=5$ versus 4.06\% for REVS, increasing to 53.34\% at $B=20$. In contrast, broad changes after independent edits are small on random contexts, WikiText, and downstream tasks; for OVE-MA at $B=5$, task changes are below 0.4 percentage points in magnitude (Table~\ref{tab:cross_model_locality}). Thus, broad probes can obscure effects on other contexts requiring the edited tokens.

\paragraph{Cumulative editing.}
Through 50 retained edits, OVE-MA maintains strong greedy mitigation with substantially less broad degradation than at 100 edits (Table~\ref{tab:cumulative_results}). At $B=5$, 92.0\% of requests remain mitigated with $\Delta$PPL $+1.30$ and task changes of at most 2.2 percentage points; at $B=20$, mitigation is 100\% with $\Delta$PPL $+3.21$ and task changes of at most 5.2 points. Full REVS mitigates 88.0\% with $\Delta$PPL $+0.98$.

\begin{table}[t]
\centering
\small
\setlength{\tabcolsep}{1.7pt}
\begin{tabular}{@{}llrrrrrr@{}}
\toprule
\multirow{2}{*}{Method} &
\multirow{2}{*}{$N$} &
\multirow{2}{*}{\shortstack{Greedy\\mitig. $\uparrow$}} &
\multirow{2}{*}{Rec. $\downarrow$} &
\multirow{2}{*}{$\Delta$PPL $\downarrow$} &
\multicolumn{3}{c}{\(\Delta\)Tasks} \\
\cmidrule(lr){6-8}
& & & & & HS & ARC-C & QA \\
\midrule

\shortstack[l]{OVE-MA\\$B=5$}
& 10  & 86.7 & --  & +0.27  & -0.5  & +0.0 & +0.2 \\
& 25  & 89.3 & --  & +0.93  & -1.5  & -1.5 & -1.0 \\
& 50  & 92.0 & --  & +1.30  & -2.2  & -2.2 & -1.3 \\
& 100 & 98.0 & 3.7 & +52.88 & -14.8 & -7.0 & -15.3 \\

\midrule
\shortstack[l]{OVE-MA\\$B=20$}
& 10  & 100.0 & --  & +0.88   & -1.7 & +0.3 & -0.5 \\
& 25  & 100.0 & --  & +2.26   & -3.5 & -0.7 & -2.8 \\
& 50  & 100.0 & --  & +3.21   & -4.2 & -1.3 & -5.2 \\
& 100 & 100.0 & 0.3 & +255.17 & -20.0 & -5.0 & -16.2 \\

\midrule
\shortstack[l]{REVS\\full}
& 10  & 83.3 & --   & +0.006 & -0.3 & -0.7 & +0.0 \\
& 25  & 84.0 & --   & +0.59  & -0.3 & -0.8 & -0.8 \\
& 50  & 88.0 & --   & +0.98  & -0.7 & -1.0 & -1.2 \\
& 100 & 89.0 & 26.0 & +2.10  & -0.7 & -1.0 & -0.8 \\

\midrule
\shortstack[l]{REVS\\first 20}
& 10  & 0.0 & --    & +0.001  & +0.0 & +0.0 & +0.0 \\
& 25  & 1.3 & --    & +0.0014 & +0.0 & -0.2 & +0.0 \\
& 50  & 4.7 & --    & +0.0018 & +0.0 & -0.3 & -0.7 \\
& 100 & 5.3 & 99.3  & +0.001  & +0.0 & +0.0 & +0.7 \\

\bottomrule
\end{tabular}

\caption{Cumulative editing on OLMo-7B, averaged over three seeded orders of the same 100 development requests. Greedy mitigated is the percentage no longer reproduced under greedy decoding; Recovery (Rec.) uses the 21-output test and is reported only at $N=100$. WikiText $\Delta$PPL is relative to 16.59, and task deltas are accuracy changes in points.}
\label{tab:cumulative_results}
\vspace{-5mm}
\end{table}

At 100 edits, OVE-MA leaves only 3.7\% and 0.3\% recoverable at $B=5$ and $B=20$, respectively, compared with 26.0\% for full REVS, but degradation increases sharply. It is also strongly order-sensitive: $\Delta$PPL ranges from $+1.51$ to $+107.65$ at $B=5$ and from $+8.14$ to $+747.42$ at $B=20$, versus $+1.24$--$+2.60$ for REVS. Conversely, the untuned first-20 REVS truncation largely preserves model behavior but provides almost no suppression, leaving 99.3\% recoverable.

\subsection{Cross-model results}
\label{sec:cross_model}

We finally test whether the OLMo-7B advantage transfers to other models. It does not consistently: under the fixed REVS configuration, REVS achieves lower recovery than both OVE-MA budgets on OLMo-1B, SmolLM-360M, and Llama2-7B, with the largest differences on the two smaller models (App.~\ref{app:cross_model}). Locality is also model dependent: OVE-MA causes smaller WikiText NLL changes than REVS on Llama2-7B, whereas REVS is more local on SmolLM-360M. We therefore do not observe a method that dominates in both suppression and locality across models.

\paragraph{Practical implications.}
On OLMo-7B, $B=5$ provides suppression with few update operations, while $B=20$ trades stronger suppression for worse locality. Under cumulative editing, OVE-MA retains strong suppression with substantially smaller broad degradation through 50 edits, while longer edit sequences expose increasing and order-sensitive instability. Full REVS is substantially more stable at $N=100$ but leaves more continuations recoverable. The balance between suppression and locality is model- and budget-dependent.

\section{Analysis}
\label{sec:analysis}

\subsection{Why suppression varies across tokens}
\label{sec:realization_analysis}

We examine why the same suppression rule varies across target tokens. The realization fraction $c_t$ measures how much of the requested change in the target token's output projection is actually achieved by the least-squares update (Eq.~\ref{eq:realization}). Holding the neuron activation fixed, suppression scales an initially positive target contribution by $1-(1+\beta)c_t$. With $\beta=20$, this contribution changes sign only when $c_t > 1/21 \approx 0.0476$. Thus, even with identical hyperparameters, the operator can strongly suppress one token while only weakly affecting another.
The empirical results are consistent with this prediction. Among edited single-anchor requests, all 28 requests with median realization below 0.05 remain recoverable under the full 21-output test, whereas recovery falls to 43.36\% among the 256 requests with median realization of at least 0.20.

Realization therefore explains part of OVE's variation across requests and motivates selecting anchors by predicted editability. This association is observational because tokens, contexts, activations, and selected neurons differ between groups.

\subsection{Failures depend on the intervention}
\label{sec:failure_analysis}

Different editing configurations fail on different continuations. Across EOS-MA, Next-best-MA, Suppress-MA at $B=5$, and full REVS, at least one configuration prevents recovery for 97\% of the OLMo-7B pairs, and each configuration succeeds on some requests where the others fail. This is a retrospective oracle comparison, not a tested selection strategy, but it suggests that per-request choice of intervention could improve over any fixed configuration (App.~\ref{app:modes}). Failures also change with editing strength. Of 19 pairs recoverable under all four configurations, Suppress-MA at
$B=20$ prevents recovery for 17, while most residual failures are different examples. Failures are therefore configuration- and budget-dependent, but stronger editing also increases locality damage.

\section{Conclusion}
\label{sec:conclusion}

We introduced OVE, a post-hoc method for suppressing reproduced continuations by modifying localized MLP output vectors rather than removing their entire contribution. OVE-MA uses the realizability of updates to adaptively select continuation positions under a fixed update budget.

On OLMo-7B, OVE-MA achieves lower recovery than REVS with substantially fewer output-vector updates: at $B=5$, it leaves 24.45\% of 634 validated continuations recoverable, compared with 32.02\% for full REVS, using $\sim$75-fold fewer updates. This advantage is not universal, however. OVE-MA causes greater same-target damage, long cumulative edit sequences can degrade broader model performance, and its recovery advantage does not transfer consistently across models.

Our analyses further show that suppression depends strongly on which continuation positions are edited and on how well the requested projection change can be realized. Different editing configurations also exhibit complementary failure modes, suggesting that future methods may benefit from deciding not only how strongly to edit, but where and which intervention to apply.

More broadly, our results show that evaluating memorization mitigation requires more than measuring greedy reproduction. Recovery should be tested under multiple decoding procedures and jointly with locality and cumulative side effects. OVE reduces reproduction under the tested conditions, but this does not establish removal of the underlying information.

\section*{Limitations}

\paragraph{Scope of the task and recovery evaluation.}
OVE is a white-box intervention that requires access to model weights and a known reproduced prefix--continuation pair. We study reproduction of fixed 20-token continuations rather than broader forms of memorization or unlearning. Recovery is evaluated with a finite set of decoding procedures; different prompts, longer generations, or more extensive search may recover continuations that our tests miss. Moreover, BLEU-based recovery measures similarity to the reference continuation and does not capture paraphrased expression of the same information. Our results therefore establish reduced reproduction under the tested conditions, not removal of the underlying information.

\paragraph{Data and model generalization.}
OLMo-7B is our primary development and evaluation setting because its weights and training corpus are public. The 634-pair evaluation includes 240 development pairs and examples inspected during exploratory experiments, so it should not be interpreted as an untouched test set. Although we additionally evaluate OLMo-1B, SmolLM-360M, and Llama2-7B, the relative performance of OVE-MA and REVS varies substantially across models. The Llama2-7B evaluation set is mined from SlimPajama because the model's exact training corpus is unavailable; these experiments therefore establish reproduction of proxy-corpus text rather than training-set membership. Broader evaluation on independently held-out examples, larger models, and additional model families is needed to determine when OVE's advantages transfer.

\paragraph{Baseline comparability.}
OVE-MA and REVS differ not only in their output-vector updates, but also in localization and continuation-position selection. Their end-to-end comparison therefore does not isolate which component accounts for differences in recovery or locality. In addition, our main REVS experiments use fixed rank-based settings derived from the published method rather than systematic model-specific tuning. Alternative vocabulary-relative thresholds change the Llama2-7B results (App.~\ref{app:ove_revs}), illustrating that conclusions can depend on baseline adaptation. A controlled comparison that exchanges localization, position-selection, and editing components between the methods would better isolate their individual contributions.

\paragraph{Locality and cumulative editing.}
Our locality tests cover several complementary probes but cannot establish that an edit is local in general. Same-target controls test only contexts sharing the edited target tokens and are restricted to predictions the unedited model originally gets correct. Random-context and downstream-task evaluations use fixed, relatively small panels. Cumulative editing is evaluated only on OLMo-7B using three seeded orders of the same 100 development requests; these runs are therefore not independent. Consequently, the observed degradation does not characterize longer editing sequences, other task distributions, or other models.

\paragraph{Update counts and edit magnitude.}
Our comparison counts output-vector update operations, including repeated updates to the same vector. This count does not constrain update magnitude or total parameter displacement. Indeed, fewer update operations need not imply a smaller aggregate parameter change (App.~\ref{app:cumulative}). Likewise, the first-five and first-twenty REVS results are truncations of its existing update sequence rather than variants optimized for those budgets.

\section*{Ethical Considerations}

OVE is intended to reduce reproduction of potentially sensitive or copyrighted text. Because weight edits can also affect unrelated model behavior, deploying such interventions without locality evaluation could introduce unintended changes. The same editing capability could also be used to selectively suppress model outputs; our experiments therefore motivate evaluating both suppression and collateral effects before deployment.

\paragraph{Use of generative tools.}
We used LLM-based tools to assist with literature retrieval, language editing and as coding assistants during experiment implementation and debugging. They were not used to formulate the research questions or experimental design. All generated content and code used in the paper were reviewed and verified by the authors.

\newpage

{
\small
\bibliography{references}
}


\newpage

\appendix

\section{Mining and validation}
\label{app:mining}

\subsection{OLMo-7B candidates}

We scan 5M sentences from each of the Books and CC-En-Head
partitions of Dolma v1.7. Shards are read in sorted order.
We split documents with NLTK's Punkt tokenizer and sample
sentences using seed $42+i$ for document index $i$.

Teacher forcing provides next-token predictions for all positions in a sentence in one forward pass. We compare these predictions with the reference in every contiguous 20-token window and retain the window with the highest BLEU if it exceeds 0.5. The prefix ends at the start of that window; the following 20 reference tokens form the candidate continuation.

The original mining records contain the sentence index, prefix,
window position, and teacher-forced BLEU. We recover the reference
continuation by replaying the sentence stream. For all 6{,}831
deduplicated OLMo-7B candidates, the recovered prefix matches the
stored prefix and at least 20 reference tokens remain. We tokenize
without adding a beginning-of-sequence token and retain token IDs
for editing and evaluation.

Both screening and autoregressive validation use sentence-level
BLEU on token IDs, with uniform 1--4-gram weights and no smoothing.
Validation generates 20 tokens greedily from the prefix alone and
scores the output before its first EOS. We retain pairs with
AR-BLEU $\geq 0.75$. Unlike the teacher-forced threshold, this
stricter criterion determines inclusion in the evaluation pool.

After deduplication, Books contributes 2{,}049 candidates and 333
validated pairs; CC-En-Head contributes 4{,}782 candidates and 301
validated pairs. Table~\ref{tab:appendix_data} summarizes mining
and evaluation counts across models.

\begin{table*}[tp]
\centering
\small
\setlength{\tabcolsep}{5pt}
\begin{tabular}{@{}lcccc@{}}
\toprule
Evaluation set &
Candidates &
Validated &
\shortstack{Validated (\%)} &
Evaluated \\
\midrule
OLMo-7B            & 6{,}831   & 634   & 9.28 & 634 \\
OLMo-1B            & 4{,}968   & 206   & 4.15 & 205 \\
SmolLM-360M        & 18{,}051  & 776   & 4.30 & 300 \\
Llama2-7B          & 130{,}413 & 7{,}132 & 5.47 & 300 \\

\bottomrule
\end{tabular}
\caption{Candidate mining and evaluation counts. Validation requires greedy generation from the prefix to reach AR-BLEU $\geq0.75$. The relatively low validation rates show that teacher-forced screening identifies many candidates that are not reproduced autoregressively.}
\label{tab:appendix_data}
\label{tab:mining}
\label{tab:cross_model_data}
\end{table*}

Of the 6{,}831 OLMo-7B candidates, 634 (9.28\%) satisfy the
autoregressive criterion and 336 reproduce all 20 reference tokens
exactly under greedy decoding. Across all candidates, median
AR-BLEU is 0.148 and median exact-prefix length is two tokens.
Teacher-forced and autoregressive BLEU have Spearman correlation
$\rho=0.409$. Teacher-forced mining therefore provides an efficient
candidate screen, but does not establish reproduction from the
prefix alone.

\subsection{Development and evaluation sets}
\label{app:data_splits}

We first select 140 development pairs, stratified by source,
exact reproduction, AR-BLEU, and the number of localized neurons
(seed 20260909). We then select 100 additional pairs from the
remaining 494, also stratifying by whether the memorization anchor
skips an initial token (seed 20260910). Their union forms the
240-pair development set, leaving 394 pairs.

An exploratory recovery analysis examined 160 of those 394 pairs,
including all 32 single-anchor localization failures. The other
128 were stratified by post-edit greedy recovery, measured
realization fraction, and pre-edit exact reproduction. These
analyses informed subsequent method development. The combined
634-pair evaluation is therefore not an untouched test set.

The completed comparison evaluates all 634 pairs under the same
recovery protocol, including the anchor policies, edit modes,
mechanism controls, and truncated REVS variants. Performance probes
use a fixed panel of 100 editing requests. Earlier development
comparisons are not mixed into these completed results.
Cross-model sets are described in
App.~\ref{app:cross_model_models}.

\subsection{Model loading and numerical details}

We load models with TransformerLens
\citep{nanda2022transformerlens}, enabling
\texttt{center\_unembed} and \texttt{center\_writing\_weights}
and disabling automatic BOS insertion. The formulas use this
parameterization: the unembedding is centered over vocabulary
coordinates and each output-vector row is centered over its
hidden coordinates. Model weights and forward passes use fp32.
For gated MLPs, $a_i$ is read from \texttt{hook\_post} after the
gate product and before the output projection.

OLMo-7B has 32 layers, hidden dimension 4{,}096, and 11{,}008
output vectors per MLP, giving 352{,}256 vectors in total.
Both OLMo models use LayerNorm and have 50{,}304 unembedding
columns but 50{,}280 tokenizer entries. SmolLM-360M and
Llama2-7B use RMSNorm, with 49{,}152 and 32{,}000 vocabulary
entries, respectively. The least-squares objective uses all
unembedding columns, including OLMo's 24 additional columns.
Architecture dimensions are included in the accompanying
model-configuration export.

\paragraph{Distance computation and normalization.}
We use $\epsilon_{\mathrm{dist}}=10^{-6}$ and
$\epsilon_{\mathrm{score}}=10^{-5}$ in
Equation~\ref{eq:neuron_score}. The distance offset is added
to every coordinate before taking absolute values, following
PyTorch's pairwise-distance implementation. The implementation
divides each score by the sum of the neuron scores and the
incoming residual contribution's score, computed using the
same rule. This shared positive denominator preserves the
within-layer ranking in exact arithmetic. Scores at the
floor can tie; no stable tie order is imposed.

\paragraph{Least-squares solver.}
We cache an fp64 QR factorization
$\mathbf{W}_U^\top=\mathbf{Q}\mathbf{R}$. For a requested
vocabulary change $\delta_i\mathbf{e}_s$, we solve
\begin{equation}
\mathbf{R}\Delta\mathbf{v}_i
=\mathbf{Q}^\top(\delta_i\mathbf{e}_s)
\end{equation}
by triangular substitution. This avoids forming an explicit
pseudoinverse. We cast the edited vector to the model's
weight dtype before writing it and do not re-center the
update. Direct projection calculations omit the final
normalization and unembedding bias.

\paragraph{Settings.}
Localization retains at most five MLP layers with layer
importance above 0.2, takes $k=5$ neurons per layer, and
applies the top-/bottom-50 vocabulary filter. We use
$\beta=20$ and OVE-MA budgets $B\in\{5,20\}$. The strength
was selected through development experiments on the initial
140 pairs and subsequent refinement on the additional 100.
The localization thresholds were carried forward from the
original pipeline. These settings are reused across models.

\paragraph{Reproducibility.}
The completed runs use Python 3.10.20, PyTorch 2.10.0 with CUDA 12.8, TransformerLens 3.0.0b3, and Transformers 5.3.0 on NVIDIA RTX A6000 GPUs. Across the project, we identify approximately 3,300 GPU-hours of compute, including failed runs, resumes, hyperparameter selection, and exploratory experiments not included in the paper.

\subsection{Anchor eligibility}
\label{app:stopword_filter}

The exclusion set contains the first token ID of each NLTK English stopword in lowercase and capitalized form, with and without a leading space; punctuation with and without a leading space; and single ASCII letters with and without a leading space. The punctuation list is \texttt{! . , ? : ; ( ) " ' -} and the em dash. Deduplication gives 650 token IDs for OLMo-7B. The set is constructed with each model's tokenizer. As a token-ID filter, it can also exclude a token that begins a longer word.

Eligibility is checked on the reference continuation. At position $j$, the edit prefix consists of $x$ followed by $y_{<j}$, and the target is $y_j$. No generated tokens extend this prefix. For 363 of the 634 OLMo-7B pairs, the first eligible target is $y_0$; the others require one to five reference tokens before it. Recovery is always evaluated from the original prefix $\mathbf{x}$.

\subsection{Anchor scoring}
\label{app:anchor_scoring}

\paragraph{Rarity.}
Rarity is $-\log((n_t+1)/(N+|V|))$, where $n_t$ is the
target token's count in a reference pool of 30{,}011 Dolma
sentences and $N=902{,}531$ is the total token count.
App.~\ref{app:locality_protocol} describes this pool.

\paragraph{Redirection score.}
For redirection modes, we use
\begin{equation}
R_j=\sum_{i\in\mathcal{N}_j}
a_i(\mathbf{u}_s-\mathbf{u}_{y_j})^\top
\Delta\mathbf{v}_i.
\label{eq:redirect_anchor_score}
\end{equation}
This estimates the increase in the substitute--target
projection margin, holding activations fixed. The editability
score $E_j$ uses the same update-norm denominator as in
Equation~\ref{eq:editability}. Scores describe direct
projections, not changes in final token probabilities.

\subsection{Localization coverage}
\label{app:neuron_filter}

At the memorization anchor, 585 of the 634 OLMo-7B pairs
receive an edit. All 49 failures occur because no candidate
survives the vocabulary filter; every request has an MLP
layer above the layer threshold. OVE-MA finds an editable
anchor for all 634 pairs. These are per-request coverage
measurements, not candidate-neuron rejection rates. They do
not establish that the filter retains every relevant neuron.

\section{Baseline implementations}
\label{app:baseline_implementation}

\subsection{REVS}
\label{app:ove_revs}

Our REVS implementation follows the official code
\citep{ashuach2025revs}. It edits one target at the memorization anchor, using the same prefix
as single-anchor OVE.
This adapts the original selection of two rare tokens from a
sensitive span. Table~\ref{tab:revs_settings} lists the fixed
settings used across models. Ranks are zero-based: the number of
vocabulary entries with a strictly larger value.

\begin{table}[tbp]
\centering
\small
\setlength{\tabcolsep}{3pt}
\begin{tabular}{@{}lr@{}}
\toprule
Setting & Value \\
\midrule
Activation candidates per layer & 100 \\
Maximum neurons per layer & 30 \\
Initial neuron count & 5 \\
Neuron rank window & [30{,}000, 45{,}000] \\
MLP rank window & [1{,}000, 45{,}000] \\
Residual rank window & [1{,}200, 4{,}000] \\
Initial coordinate change & $-10$ \\
Growth / shrink factors & 1.3 / 0.8 \\
Iterations per search & 100 \\
\bottomrule
\end{tabular}
\caption{Fixed REVS settings. Neuron counts and neuron/residual
rank windows follow the published GPT-J configuration; the MLP
window and iteration constants follow the official implementation.}
\label{tab:revs_settings}
\end{table}

REVS visits layers in order and runs a new forward pass after each
layer's edits. It selects candidates by signed activation and then
by the target's rank in their vocabulary projections. Projection
and reconstruction include the final normalization and use an
explicit fp32 pseudoinverse. LayerNorm and RMSNorm use their
respective forward and inverse operations; the Llama2 and SmolLM
results use the corrected RMSNorm implementation.

\paragraph{Truncation diagnostics.}
For the first-five and first-twenty comparisons, we compute the
full update sequence, restore the weights, and replay only its first
five or twenty operations in their original order. These are
truncations of the fixed REVS configuration, without retuning for
the smaller budgets. Both are evaluated on all 634 OLMo-7B pairs.
The cumulative first-twenty arm uses the same rule at each request.

Both truncations leave all 634 pairs recoverable under the
21-output test. Their greedy residual rates are 99.53\% and
97.79\%, respectively. Thus, the original REVS update sequence
cannot simply be truncated to OVE-sized budgets while retaining
its suppression effectiveness.

\paragraph{Llama2 settings and precision.}
The fixed neuron and MLP rank windows extend beyond Llama2's
32{,}000-entry vocabulary. We therefore report a separate configuration with vocabulary-relative thresholds on the Llama2-7B evaluation set (Table~\ref{tab:cross_model_updates}). These comparisons vary
the rank thresholds without otherwise constituting systematic
model-specific tuning.


The fp32 projection-and-reconstruction check has a maximum relative
error of $4.27\times10^{-3}$ on Llama2, exceeding the upstream
$10^{-3}$ tolerance; the implementation uses a $10^{-2}$ assertion
threshold. A separate 100-pair precision check keeps the rank
windows fixed and recomputes the pseudoinverse and hidden-state
inversion in fp64. The reported round-trip error falls to
$1.97\times10^{-4}$, while recovery remains 20.00\% in both
cases. This check varies precision separately from the
rank-threshold configuration.

\subsection{Mechanism controls}
\label{app:mechanism_controls}

\paragraph{Single-anchor controls.}
Zero ablation sets the vectors selected by single-anchor OVE to
zero, removing their contributions at every position and on every
input. We evaluate it on every request of each model's evaluation
set. On OLMo-7B, random-neuron editing matches the single-anchor
update count, sampling layer--neuron pairs with replacement from
its selected layers and applying the same suppression rule without
the vocabulary filter. Repeated draws count as separate operations.
The OLMo-7B comparisons use all 634 pairs.

\paragraph{Multi-anchor controls.}
At $B=5$, we use the recorded Suppress-MA sequence of selected anchors and layers. The zero control sets the selected vectors to zero. The random control substitutes random neurons and applies suppression while matching the recorded operation count. These controls condition on OVE-MA's selections; they do not independently optimize an anchor policy for ablation or random editing. Table~\ref{tab:mechanism_controls} reports their recovery results; additional locality measurements appear in App.~\ref{app:cross_model}.

\paragraph{Activation steering.}
On OLMo-7B, the direction is the difference between mean residual
activations for the 240 development prefixes and 500 WikiText-103
validation passages. The control passages contain 31--499 stripped
characters and are sampled after shuffling with seed 42. We read
the last-position residual after block 25 and subtract half this
direction ($\alpha=0.5$) during generation. Steering changes no
weights but requires an inference-time intervention. Its reported
vector-update count is therefore zero.

\begin{table}[t]
\centering
\small
\setlength{\tabcolsep}{3pt}
\begin{tabular}{@{}lrrr@{}}
\toprule
Control &
\shortstack{Greedy\\residual $\downarrow$} &
Recovery $\downarrow$ &
\shortstack{Mean\\updates} \\
\midrule
Zero, single anchor   & 68.61 & 86.91  & 2.81 \\
Random, single anchor & 99.68 & 99.84  & 2.81 \\
Zero, MA selections   & 49.84 & 71.92  & 4.99 \\
Random, MA selections & 99.53 & 100.00 & 4.99 \\
Activation steering   & 13.88 & 28.39  & 0 \\
\bottomrule
\end{tabular}
\caption{Mechanism controls on all 634 OLMo-7B pairs.
Recovery metrics are percentages. Single-anchor controls use the
neurons selected by OVE; MA controls follow the recorded selections
of OVE-MA at $B=5$. Activation steering changes no weights, so its
output-vector update count is zero.}
\label{tab:mechanism_controls}
\end{table}


\section{Evaluation details}
\label{app:evaluation_details}

\subsection{Recovery tests}
\label{app:recovery_protocol}

All recovery tests generate from the original prefix and compare
outputs with the reference continuation using token-ID BLEU with
uniform 1--4-gram weights and no smoothing. Only tokens before the
first EOS are scored; an empty output scores zero. Recovery requires
at least one tested output with AR-BLEU $\geq0.75$. Exact-20
recovery requires all 20 reference tokens to match without EOS.

The main test includes one greedy output, all five beams from
width-five beam search, and five samples under each of three
conditions: $T=0.7$, $T=1.0$, and $T=1.0$ with top-$p=0.9$.
Each output has a 20-token horizon. Beam search uses cumulative
log probability without length normalization. Completed beams
retain their scores and are padded with EOS. Greedy and sampling
generation use a key--value cache; beam search recomputes the
sequence.

Requests without a weight edit remain in the denominator and reuse
the unedited model's recovery outputs. Their validated greedy
output already meets the AR-BLEU threshold. Exact-20 recovery is
measured from the outputs, not inferred from the absence of an
edit. Failed or missing runs are not counted as successful
suppression.

\paragraph{EOS and repetition.}
Raw greedy generations are retained for 20 steps even after EOS.
EOS truncation records whether EOS occurs within this horizon.
Repetitive degeneration records whether one token ID occurs at
least ten times in the raw generated sequence. These diagnostics
are distinct from BLEU, which stops before the first EOS. Every
primary arm has a stored greedy output for every request.

\paragraph{Additional decoding attempts.}
On a fixed panel of 100 OLMo-7B requests, a sensitivity analysis
retains the original 21 outputs and adds ten width-ten beams and
fifteen further samples per sampling condition. This gives 76
outputs, so recovery cannot decrease. Table~\ref{tab:decoder_sensitivity}
shows how much additional recovery is exposed by the larger test.

\begin{table}[tbp]
\centering
\small
\setlength{\tabcolsep}{4pt}
\begin{tabular}{@{}lrrr@{}}
\toprule
Method &
21 outputs &
76 outputs &
$\Delta$ recovery \\
\midrule
OVE, single anchor & 61.00 & 69.00 & +8.00 \\
OVE-MA, $B=5$      & 25.00 & 29.00 & +4.00 \\
OVE-MA, $B=20$     & 1.00  & 2.00  & +1.00 \\
REVS, full         & 28.00 & 42.00 & \textbf{+14.00} \\
\bottomrule
\end{tabular}
\caption{Recovery (\%) on the same 100 OLMo-7B requests as the
number of tested continuations increases. $\Delta$ recovery is the
increase from 21 to 76 outputs. Additional decoding increases
observed recovery for every method, with the largest increase for
REVS. These subset results are separate from the 634-request
comparison.}
\label{tab:decoder_sensitivity}
\end{table}

The expanded test remains finite and therefore cannot establish
that a continuation is unrecoverable by other decoding procedures.

\subsection{Locality controls}
\label{app:locality_protocol}

The OLMo-7B reference pool contains 30{,}011 Dolma sentences.
We exclude mined sentences and documents containing them, retain
sentences of at least 20 tokens, and truncate longer sentences to
40 tokens. We retain at most 30 sentences per Books document or
five per CC-En-Head document, using seed 42. Prediction positions
begin after at least eight prefix tokens. Cross-model control pools
use up to 4{,}000 mined sentences that fail autoregressive
validation for the corresponding model.

\paragraph{Shared same-target controls.}
For each request, we form the union of anchor targets used by the
compared methods and the first eligible target. For each target,
we sample up to 40 candidate prediction positions and retain up to
20 with the highest unedited target probability, without an
additional probability cutoff. We exclude pool sentences sharing
a six-token sequence with the request's prefix and continuation.
The resulting panel is frozen and shared across methods.

Only controls whose true next token is predicted correctly by the
unedited model are eligible. Same-target damage is the fraction of
these predictions that become incorrect. We average within each
request and then across requests with at least one eligible
control. Counts of controls refer to request--context pairs; the
same pool context can appear in more than one request's panel.

\begin{table}[tbp]
\centering
\small
\setlength{\tabcolsep}{5pt}
\begin{tabular}{@{}lrr@{}}
\toprule
Evaluation set &
\shortstack{Eligible\\requests} &
\shortstack{Eligible\\controls} \\
\midrule
OLMo-7B            & 632 & 35{,}065 \\
OLMo-1B            & 201 & 10{,}076 \\
SmolLM-360M        & 299 & 21{,}105 \\
Llama2-7B          & 299 & 16{,}856 \\
\bottomrule
\end{tabular}
\caption{Denominators for the shared same-target locality panels.
Eligible requests contain at least one control on which the unedited
model predicts the true next token correctly. Eligible controls count
request--context pairs.}
\label{tab:control_counts}
\end{table}

Damage on a method's own target set is a separate diagnostic and is
not substituted for the shared-panel comparison.

\paragraph{Random controls.}
All methods use the same 20 random prediction contexts per request.
We measure full-distribution
$\mathrm{KL}(p_{\mathrm{pre}}\|p_{\mathrm{post}})$ and average first
within requests and then across requests, using means for every
model. Unedited requests are retained.

\subsection{Performance tests}
\label{app:performance_protocol}

Each evaluation set has a fixed panel of 100 editing requests shared
across methods. After each independent request, we evaluate the same
100 WikiText-103 texts and 200 items each from HellaSwag,
ARC-Challenge, and Wikidata QA. Thus, 100 editing requests and
100 WikiText texts describe different sample dimensions. Original
weights are restored between requests.

The WikiText sample uses the first 100 passages from the project's
200-passage test file, truncates each to 256 tokens, and retains
passages with more than 16 tokens. We compute mean token NLL within
each text and average texts equally. Perplexity is the exponential
of this mean. We then average changes across the 100 editing
requests. The earlier 200-text WikiText probe is not mixed into
these results.

HellaSwag and ARC-Challenge score choices by mean continuation
log probability. Wikidata QA generates ten tokens greedily and
accepts an answer when a reference occurs in the lowercased,
alphanumeric-normalized output. Task changes are reported in
percentage points. CommonGen is excluded from the completed
performance comparison.

\subsection{Uncertainty and sampling}
\label{app:statistics}

Confidence intervals use percentile bootstrap resampling over
requests. Paired recovery comparisons resample the same request
indices for both methods, with 10{,}000 resamples and seed
20260915. Exact McNemar tests use the discordant request outcomes.
Locality and performance tables report point estimates; small
differences are not interpreted as statistically established
improvements.

The completed sampling runs use deterministic per-condition seeds
from CRC32 of the run tag, dataset, request ID, content hash, and
sampling condition. The seed definition is shared across methods
and replaces the process-dependent method-name hash used in some
earlier development runs. The three cumulative runs contain the same 100 requests in different seeded orders and are therefore not independent. Their trajectories are summarized by run rather than treating the 300 request occurrences as independent.

\section{Editing diagnostics}
\label{app:editing_diagnostics}

\subsection{Anchor-selection diagnostics}
\label{app:anchor_diagnostics}

MEM, RARE, and EDIT each permit at most five output-vector updates.
Localization and anchor selection use the unedited model. The
completed comparison evaluates all 634 OLMo-7B pairs. MEM, RARE,
and EDIT use the full five-update budget on 111, 64, and 124
requests, respectively, so a common cap does not imply an identical
realized update count. MEM also differs from uncapped single-anchor
OVE, which applies every retained update at the memorization anchor.

\begin{table}[tp]
\centering
\small
\setlength{\tabcolsep}{4pt}
\begin{tabular}{@{}lrrrr@{}}
\toprule
Policy &
\shortstack{Edited\\requests $\uparrow$} &
\shortstack{Mean\\updates} &
\shortstack{Mean\\realization $\uparrow$} &
\shortstack{Shared\\damage $\downarrow$} \\
\midrule
MEM  & 585          & 2.61 & 0.1866          & 6.92 \\
RARE & \textbf{634} & 2.35 & 0.0880          & \textbf{0.86} \\
EDIT & \textbf{634} & 2.93 & \textbf{0.2319} & 10.70 \\
\bottomrule
\end{tabular}
\caption{Anchor-selection diagnostics on all 634 OLMo-7B pairs.
Higher realization indicates that a larger fraction of the requested
projection change is achieved; lower shared damage is better.
EDIT selects anchors with the highest measured realization but also
causes the greatest same-target damage. Mean realization averages
per-request medians over edited requests.}
\label{tab:anchor_diagnostics}
\end{table}

EDIT therefore identifies positions at which the requested update is
better realized, but this greater editability comes with more
same-target damage. RARE shows the opposite pattern: it achieves
full coverage with little same-target damage, but its low realization
is consistent with its poor suppression performance in
Table~\ref{tab:anchor_selection}.

\paragraph{Development-set associations.}
A separate analysis of the 240 development pairs examined 2{,}494
eligible, localizable positions. Rarity and $c_t$ have Spearman
correlation $-0.8417$ across these positions and $-0.829$ across
their 1{,}647 distinct target tokens. Among the 223 requests with
a localizable memorization anchor, MEM and RARE select the same
position on 36, MEM and EDIT on 15, RARE and EDIT on 11, and all
three on two. These exploratory associations concern the development
set, whereas the recovery and locality comparison uses all 634
requests.

\subsection{Layer distribution of edits}
\label{app:layer_distribution}

\begin{figure*}[t]
    \centering
    \includegraphics[width=\textwidth]{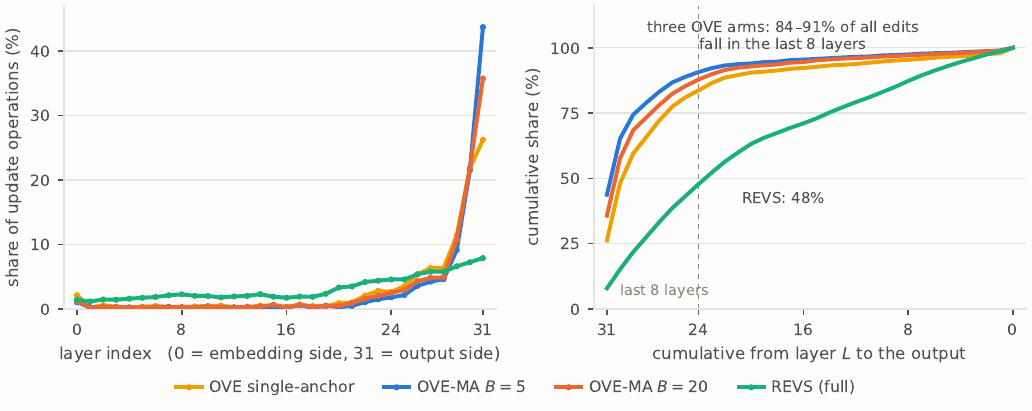}
    \caption{Distribution of output-vector update operations across OLMo-7B layers. Each curve is normalized by the total number of update operations for that method across the 634 evaluation requests. OVE variants concentrate their updates in the final layers, whereas REVS distributes them more broadly across model depth. Repeated edits to the same output vector count as separate operations, consistent with our update-budget definition.}
    \label{fig:layer_distribution}
\end{figure*}

Figure~\ref{fig:layer_distribution} shows how output-vector update
operations are distributed across OLMo-7B's 32 MLP layers. OVE
variants concentrate their edits in later layers: 83.7\% of
single-anchor OVE updates, 90.7\% of OVE-MA updates at $B=5$,
and 87.8\% at $B=20$ occur in layers 24--31. The corresponding
fraction for REVS is 47.8\%, with its updates distributed more
broadly across model depth. For OVE-MA at $B=5$, 43.7\% of all
update operations occur in the final layer alone.

Increasing the OVE-MA budget primarily adds updates within the same
late-layer region rather than shifting editing toward earlier layers.
The first and later selected anchors also have similar depth profiles:
at $B=5$, their mean layer indices are 27.9 and 28.8, respectively;
at $B=20$, both are approximately 27.7. Thus, additional anchors
primarily identify further vectors in the same late-layer region.

\section{Edit-mode ablations}
\label{app:modes}

\paragraph{Redirection modes.}
The three redirection modes use the same least-squares operator as
suppression. For each localized neuron $i$, they edit a substitute
coordinate $s$ and request
\begin{equation}
q_i=\beta\,\mathbf{u}_t^\top\mathbf{v}_i.
\label{eq:redirection}
\end{equation}
EOS uses the model's end-of-sequence token. Redact uses the first
token of \texttt{****}. Next-best uses the current model's
highest-ranked token other than $t$ at the anchor prefix. These are
names for the requested interventions; the least-squares update does
not guarantee that the substitute becomes the model's next prediction.

All modes use $k=5$ and $\beta=20$. Single-anchor modes edit all
retained vectors at the memorization anchor. Multi-anchor modes use
$B=5$ and the suppression or redirection score appropriate to their
objective. Table~\ref{tab:modes} compares all four modes.

\begin{table*}[tp]
\centering
\small
\setlength{\tabcolsep}{3pt}
\begin{tabular}{@{}lrrrrrrr@{}}
\toprule
Mode &
\shortstack{Greedy\\residual $\downarrow$} &
\shortstack{Recovery\\$\downarrow$} &
\shortstack{Repeat\\$\downarrow$} &
EOS &
\shortstack{Shared\\damage $\downarrow$} &
\shortstack{Random KL\\($\times10^{-3}$) $\downarrow$} &
\shortstack{WikiText\\$\Delta$NLL $\to0$} \\
\midrule
\multicolumn{8}{@{}l}{\textbf{Single anchor}} \\
EOS       & 51.10 & 79.97 & 1.26 & 44.95 & 2.74 & 3.856 & 0.006595 \\
Redact    & 98.58 & 99.84 & 1.74 & 0.00  & 0.16 & 0.016 & 0.000119 \\
Next-best & 45.74 & 65.46 & 5.36 & 0.00  & 3.09 & 3.875 & 0.001473 \\
Suppress  & 41.01 & 63.56 & 1.42 & 0.00  & 7.11 & 0.273 & 0.001014 \\
\midrule
\multicolumn{8}{@{}l}{\textbf{Multi-anchor, $B=5$}} \\
EOS       & 12.15 & 18.14 & 0.47 & 89.59 & 8.82  & 3.037 & 0.002862 \\
Redact    & 97.48 & 99.37 & 2.05 & 0.00  & 0.27  & 0.045 & 0.000172 \\
Next-best & 16.40 & 28.08 & 9.15 & 0.00  & 8.96  & 6.291 & 0.002707 \\
Suppress  & 14.51 & 24.45 & 0.95 & 0.00  & 19.31 & 0.432 & 0.001246 \\
\bottomrule
\end{tabular}
\caption{Edit modes on all 634 OLMo-7B requests. Greedy residual,
Recovery, Repeat, EOS, and Shared damage are percentages. Arrows
indicate the preferred direction for suppression and locality metrics;
EOS is reported as a generation diagnostic rather than an objective
to maximize or minimize. Coverage is 585/634 for every single-anchor
mode and 634/634 for every multi-anchor mode. All modes share the
same 632 eligible locality requests and 35{,}065 controls.}
\label{tab:modes}
\label{tab:ma_modes}
\end{table*}

\paragraph{Changes to generated output.}
The modes expose different ways in which recovery can decrease.
EOS occurs within the raw 20-token greedy output for 89.59\% of
EOS-MA requests, showing that much of its apparent suppression
comes with early termination. Next-best-MA has no EOS occurrences
but the highest repetitive-degeneration rate among these modes
(9.15\%). Redact-MA, by contrast, leaves 99.37\% recoverable.
Thus, low recovery alone does not imply that an edit preserves
normal generation behavior.

\paragraph{Effects on other inputs.}
Suppress-MA has greater same-target damage on the shared panel than
the other $B=5$ modes, but smaller random-control KL and WikiText
NLL increases than EOS-MA and Next-best-MA. These probes therefore
capture distinct side effects: damage to other contexts requiring
the edited tokens, broad distributional changes, and changes in
held-out language-model likelihood.

\paragraph{Complementarity across configurations.}
We use a retrospective union analysis to test whether different configurations succeed and fail on the same continuations. At least one of the four single-anchor OVE modes prevents recovery for 312 of 634 pairs (49.21\%). For EOS-MA, Next-best-MA, and Suppress-MA at $B=5$, together with full REVS, the corresponding retrospective union is 615 pairs (97.00\%). Their unique successes number 17, 10, 8, and 17, respectively.

These outcomes come from independently edited model states and select a successful configuration only after observing recovery. They therefore show complementary failure modes rather than the performance of a tested method-selection policy. The union also imposes no locality constraint and includes success through EOS termination. Predicting which intervention is appropriate for a given request without relying on post-edit recovery remains an open problem.

\section{Cross-model details}
\label{app:cross_model}

\subsection{Models and mining}
\label{app:cross_model_models}

We use Dolma for OLMo-1B, Cosmopedia v2~\citep{benallal2024cosmopedia} and FineWeb-Edu~\citep{penedo2024fineweb} for
SmolLM-360M, and SlimPajama for Llama2-7B. Llama2 results measure
reproduction of proxy-corpus text; membership in the model's
training data is unverified. All models use the same teacher-forced
candidate threshold and autoregressive validation criterion.
Their reference continuations were stored during mining, so
sentence-stream replay is needed only for OLMo-7B.

The recorded scan budgets are 5M Books and 9M CC-En-Head sentences
for OLMo-1B, 5M sentences from each SmolLM source, and 1.65M Book
and 9M CommonCrawl sentences for Llama2.
Table~\ref{tab:appendix_data} gives candidate, validation, and
evaluation counts.

The original selection manifest records validated parent pools of
7,131 Llama2-7B, 205 OLMo-1B, and 770 SmolLM-360M pairs,
whereas recounting the validation files yields 7,132, 206, and 776
distinct IDs, respectively. These small discrepancies do not affect
the selected evaluation sets: the evaluation IDs remain the same
300 Llama2-7B, 205 OLMo-1B, and 300 SmolLM-360M requests.
The origin of the count differences is not recorded.

\subsection{Coverage and update counts}

We transfer $k=5$, $\beta=20$, localization thresholds, and the
anchor-selection rule without model-specific tuning.
Table~\ref{tab:cross_model_updates} reports the completed
cross-model comparisons. Update means include unedited requests.
The same nominal budget can yield different operation counts because
localization and the number of anchors with positive scores vary
across models.

\begin{table*}[tp]
\centering
\small
\setlength{\tabcolsep}{3pt}
\begin{tabular}{@{}lrrrrr@{}}
\toprule
Method &
Coverage $\uparrow$ &
Updates &
\shortstack{Greedy\\residual $\downarrow$} &
Recovery $\downarrow$ &
Exact-20 $\downarrow$ \\
\midrule
\multicolumn{6}{@{}l}{\textbf{OLMo-1B} ($n=205$)} \\
OVE, single anchor & 57.07 & 1.08 & 92.20 & 98.54 & 57.07 \\
OVE-MA, $B=5$      & 98.54 & 4.68 & 70.24 & 90.24 & 48.29 \\
OVE-MA, $B=20$     & 98.54 & 10.70 & 58.54 & 84.39 & 41.95 \\
REVS, full         & \textbf{99.51} & 202.05 &
\textbf{18.05} & \textbf{36.10} & \textbf{5.85} \\
Zero, single anchor & 57.07 & 1.08 & 88.29 & 97.56 & 55.12 \\
\midrule
\multicolumn{6}{@{}l}{\textbf{SmolLM-360M} ($n=300$)} \\
OVE, single anchor & 52.67 & 1.21 & 90.33 & 92.67 & 58.67 \\
OVE-MA, $B=5$      & 99.67 & 4.67 & 65.33 & 72.67 & 43.00 \\
OVE-MA, $B=20$     & 99.67 & 14.00 & 57.00 & 65.33 & 39.00 \\
REVS, full         & \textbf{100.00} & 240.58 &
\textbf{7.00} & \textbf{14.00} & \textbf{0.33} \\
Zero, single anchor & 52.67 & 1.21 & 91.00 & 97.00 & 64.00 \\
\midrule
\multicolumn{6}{@{}l}{\textbf{Llama2-7B} ($n=300$)} \\
OVE, single anchor & 94.00 & 3.61 & 62.67 & 82.33 & 43.67 \\
OVE-MA, $B=5$      & \textbf{100.00} & 4.99 & 43.67 & 57.67 & 21.67 \\
OVE-MA, $B=20$     & \textbf{100.00} & 19.94 & 10.67 & 18.00 & 4.67 \\
REVS, full         & \textbf{100.00} & 241.67 &
\textbf{5.67} & \textbf{14.00} & \textbf{0.00} \\
Zero, single anchor & 94.00 & 3.61 & 85.00 & 93.33 & 56.00 \\
REVS, relative ranks & 100.00 & 351.52 & 10.33 & 21.33 & 3.00 \\
\bottomrule
\end{tabular}
\caption{Full recovery comparisons on the additional evaluation
sets. Coverage and recovery columns are percentages. Arrows indicate
the preferred direction. Bold marks the best effectiveness value
among the primary OVE variants and full REVS; zero ablation and
vocabulary-relative REVS are diagnostic variants. Updates average
output-vector operations over all requests, including unedited ones.}
\label{tab:cross_model_updates}
\label{tab:cross_model_full}
\label{tab:llama_holdout}
\label{tab:revs_portability}
\end{table*}

The central cross-model trend differs from OLMo-7B: under the fixed
REVS configuration, full REVS achieves lower recovery than OVE-MA
on each additional evaluation set. OVE-MA nevertheless uses far
fewer update operations, so the relative trade-off between
suppression and editing operations is model dependent.

\subsection{Locality and performance}

Table~\ref{tab:cross_model_locality} compares the primary editing
methods using the shared controls and fixed probe panels described
in App.~\ref{app:locality_protocol}. Same-target damage and random
KL use the full evaluation sets; the task and WikiText probes use
100 editing requests per set.

\begin{table*}[tp]
\centering
\small
\setlength{\tabcolsep}{2.7pt}
\begin{tabular}{@{}lrrrrrr@{}}
\toprule
Method &
\shortstack{Shared damage\\(\%) $\downarrow$} &
\shortstack{Random KL\\($\times10^{-3}$) $\downarrow$} &
\shortstack{WikiText\\$\Delta$NLL $\to0$} &
\shortstack{HellaSwag\\$\Delta$Acc $\to0$} &
\shortstack{ARC-C\\$\Delta$Acc $\to0$} &
\shortstack{QA\\$\Delta$Acc $\to0$} \\
\midrule
\multicolumn{7}{@{}l}{\textbf{OLMo-7B}} \\
OVE, single anchor & 7.11 & 0.273 & 0.001014 & -0.05 & -0.16 & +0.02 \\
OVE-MA, $B=5$      & 19.31 & 0.432 & 0.001246 & -0.17 & -0.34 & +0.02 \\
OVE-MA, $B=20$     & 53.34 & 2.321 & 0.005129 & -0.37 & -0.55 & -0.14 \\
REVS, full         & 4.06 & 0.079 & 0.002035 & -0.04 & -0.20 & +0.05 \\
\midrule
\multicolumn{7}{@{}l}{\textbf{OLMo-1B}} \\
OVE, single anchor & 0.54 & 0.027 & -0.000008 & -0.01 & -0.01 & +0.11 \\
OVE-MA, $B=5$      & 3.68 & 0.166 & 0.000056 & -0.05 & -0.04 & +0.33 \\
OVE-MA, $B=20$     & 6.04 & 0.360 & 0.000241 & -0.07 & -0.07 & +0.38 \\
REVS, full         & 9.18 & 2.159 & 0.001331 & -0.11 & -0.21 & +0.60 \\
\midrule
\multicolumn{7}{@{}l}{\textbf{SmolLM-360M}} \\
OVE, single anchor & 1.29 & 5.817 & 0.005141 & +0.13 & +0.05 & +0.21 \\
OVE-MA, $B=5$      & 5.54 & 50.757 & 0.044697 & +0.47 & +0.43 & +0.26 \\
OVE-MA, $B=20$     & 8.23 & 93.071 & 0.101492 & +0.64 & +0.51 & -0.12 \\
REVS, full         & 5.57 & 14.231 & 0.010691 & -0.13 & -0.34 & +0.55 \\
\midrule
\multicolumn{7}{@{}l}{\textbf{Llama2-7B}} \\
OVE, single anchor & 1.77 & 2.283 & 0.001171 & -0.01 & -0.03 & +0.07 \\
OVE-MA, $B=5$      & 4.28 & 2.417 & -0.000019 & +0.00 & -0.04 & +0.04 \\
OVE-MA, $B=20$     & 15.96 & 14.839 & 0.000662 & +0.01 & -0.15 & +0.04 \\
REVS, full         & 12.38 & 1061.913 & 0.873747 & -5.44 & -4.10 & -8.77 \\
\bottomrule
\end{tabular}
\caption{Locality and performance for the primary editing methods.
Shared damage and random KL use the full evaluation sets; WikiText
and task probes use 100 editing requests per set. Arrows indicate
that locality corresponds to smaller changes, not necessarily larger
task scores. Task columns are accuracy changes in percentage points.
The mixed pattern across models and metrics does not support a
universal locality advantage for either method.}
\label{tab:cross_model_locality}
\end{table*}

The locality comparison is metric- and model-dependent. On
OLMo-7B, OVE-MA causes substantially more same-target damage than
REVS despite similarly small broad-distribution changes after
independent edits. On Llama2-7B, OVE-MA produces much smaller
random-control and WikiText changes than REVS, whereas on SmolLM
REVS is more local on those broad probes. No method therefore
dominates across both suppression and locality.

\paragraph{Diagnostic controls and Llama2 baseline variants.}
For completeness, Table~\ref{tab:cross_model_locality_diagnostics}
reports the locality results for mechanism controls and the
vocabulary-relative REVS configurations.

\begin{table*}[tp]
\centering
\small
\setlength{\tabcolsep}{2.7pt}
\begin{tabular}{@{}lrrrrrr@{}}
\toprule
Method &
\shortstack{Shared damage\\(\%) $\downarrow$} &
\shortstack{Random KL\\($\times10^{-3}$) $\downarrow$} &
\shortstack{WikiText\\$\Delta$NLL $\to0$} &
\shortstack{HellaSwag\\$\Delta$Acc $\to0$} &
\shortstack{ARC-C\\$\Delta$Acc $\to0$} &
\shortstack{QA\\$\Delta$Acc $\to0$} \\
\midrule
\multicolumn{7}{@{}l}{\textbf{OLMo-7B controls}} \\
Zero, single anchor   & 1.66 & 0.468 & 0.000606 & -0.05 & -0.14 & +0.01 \\
Random, single anchor & 0.04 & 0.000 & 0.000000 & +0.00 & +0.00 & +0.00 \\
Zero, MA selections   & 4.65 & 1.499 & 0.003334 & -0.16 & -0.34 & -0.38 \\
Random, MA selections & 0.06 & 0.000 & -0.000002 & +0.00 & +0.00 & +0.01 \\
Activation steering   & 7.80 & 68.181 & 0.058198 & -1.00 & -1.50 & +0.00 \\
\midrule
\multicolumn{7}{@{}l}{\textbf{Additional single-anchor zero controls}} \\
OLMo-1B, Zero   & 0.66 & 2.874 & 0.002036 & -0.05 & -0.05 & +0.05 \\
SmolLM, Zero    & 0.26 & 2.938 & 0.008009 & +0.09 & -0.29 & -0.59 \\
Llama2, Zero & 0.27 & 0.434 & 0.001212 & -0.01 & +0.01 & -0.10 \\
\midrule
\multicolumn{7}{@{}l}{\textbf{Llama2 REVS sensitivity}} \\
Relative ranks & 7.37 & 694.326 & 0.557799 & -3.39 & -2.81 & -7.22 \\
\bottomrule
\end{tabular}
\caption{Locality diagnostics for mechanism controls and alternative
Llama2 REVS settings. These rows are separated from the primary-method
comparison in Table~\ref{tab:cross_model_locality} because they serve
as controls or sensitivity analyses rather than primary competing
configurations.}
\label{tab:cross_model_locality_diagnostics}
\end{table*}

\section{Cumulative editing}
\label{app:cumulative}

Each run starts from the unedited OLMo-7B model and applies 100 requests from the 240 development pairs without restoring weights. The completed artifacts identify the three runs by seeds 0, 1, and 2. All three runs contain the same 100 requests in different seeded orders and are therefore not independent. Methods share the request order within each run. OVE-MA re-localizes and selects anchors on the current, cumulatively edited model.

\begin{figure*}[t]
    \centering
    \includegraphics[width=\textwidth]{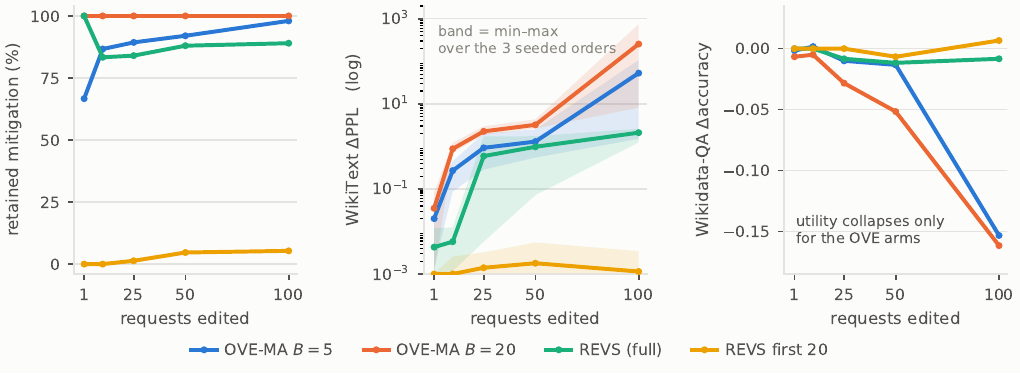}
    \caption{OLMo-7B cumulative editing. Lines show means over three runs and shaded bands show their minima and maxima, not confidence intervals. Retained mitigation is the percentage of requests edited so far that are no longer reproduced under greedy decoding; the full 21-output recovery test is performed at $N=100$ only. WikiText perplexity uses 100 fixed texts and a logarithmic vertical axis; the dotted line marks the unedited value, 16.59. The large final $B=20$ mean is driven by one run with perplexity 764.01.}
    \label{fig:cumulative}
\end{figure*}

At $N=1,10,25,50,100$ requests, we reevaluate all requests edited so far under greedy decoding and repeat the performance probes. At $N=100$, all 100 requests receive the full 21-output test. Figure~\ref{fig:cumulative} shows the retained-mitigation and perplexity trajectories.

\paragraph{Variation across runs.}
OVE-MA achieves consistently low recovery after 100 requests, but
its effect on broader model behavior varies substantially across
runs. Final WikiText perplexity ranges from 18.10 to 124.24 for
$B=5$ and from 24.73 to 764.01 for $B=20$. In contrast, full
REVS has final perplexity between 17.83 and 19.19 and recovery
between 25\% and 27\%. Thus, OVE-MA's stronger cumulative
suppression can coincide with much larger and less stable
performance degradation.

\begin{table*}[tp]
\centering
\small
\setlength{\tabcolsep}{3pt}
\begin{tabular}{@{}lrrrrrrr@{}}
\toprule
Method & Run &
\shortstack{Greedy\\residual $\downarrow$} &
\shortstack{Recovery\\$\downarrow$} &
$\Delta$PPL $\to0$ &
HellaSwag $\to0$ &
ARC-C $\to0$ &
QA $\to0$ \\
\midrule
OVE-MA, $B=5$ & 0 & 4.00 & 9.00 & +1.51   & -4.50  & -3.00  & -1.50 \\
               & 1 & 1.00 & 1.00 & +107.65 & -24.50 & -7.50  & -20.00 \\
               & 2 & 1.00 & 1.00 & +49.47  & -15.50 & -10.50 & -24.50 \\
\midrule
OVE-MA, $B=20$ & 0 & 0.00 & 1.00 & +8.14   & -13.50 & -3.50  & -11.00 \\
                & 1 & 0.00 & 0.00 & +747.42 & -38.50 & -10.50 & -27.00 \\
                & 2 & 0.00 & 0.00 & +9.94   & -8.00  & -1.00  & -10.50 \\
\midrule
REVS, full & 0 & 10.00 & 26.00 & +2.47 & -1.00 & -1.00 & -1.00 \\
           & 1 & 11.00 & 27.00 & +2.60 & -0.50 & -1.00 & -4.00 \\
           & 2 & 12.00 & 25.00 & +1.24 & -0.50 & -1.00 & +2.50 \\
\midrule
REVS, first 20 & 0 & 93.00 & 99.00  & +0.00 & +0.00 & +0.00 & +0.50 \\
               & 1 & 96.00 & 99.00  & +0.00 & +0.00 & +0.50 & +0.50 \\
               & 2 & 95.00 & 100.00 & +0.00 & +0.00 & -0.50 & +1.00 \\
\bottomrule
\end{tabular}
\caption{Individual runs after 100 cumulative requests. Greedy and
Recovery are percentages over all 100 requests. Recovery uses the
21-output test. $\Delta$PPL is relative to the unedited OLMo-7B
value of 16.59, and task columns are accuracy changes in percentage
points. Values closer to zero indicate smaller performance changes.
The table highlights substantial run-to-run variability for OVE-MA,
particularly at $B=20$.}
\label{tab:cumulative_per_seed}
\end{table*}

\paragraph{Update count does not imply smaller parameter change.}
A collision occurs when a request updates a layer--neuron pair
already touched by an earlier request. A vector counts once within
that request's collision set and can count again in a later request.
This excludes repeated operations on the same vector within a
single request.

Aggregate displacement is the square root of the sum of squared
$L_2$ changes across distinct touched vectors, measured relative
to the original model. Repeated updates are combined before
computing this norm. It therefore differs from the square root of
the sum of squared individual operation norms.

After 100 requests, mean net displacement is 13.78 for OVE-MA
at $B=5$, 24.46 at $B=20$, and 3.27 for full REVS. Thus, OVE-MA
uses far fewer output-vector update operations than REVS, but this
does not imply a smaller aggregate parameter change. The
accompanying checkpoint export records operation counts, unique
vectors, collisions, and displacement for every run.

\section{Operator attenuation and residual recovery}
\label{app:operator_diagnostics}

\subsection{Operator properties}

\paragraph{Least-squares solution.}
The normal equations for Equation~\ref{eq:objective} are
\[
\mathbf{W}_U\mathbf{W}_U^\top\Delta\mathbf{v}_i
=
\delta_i\mathbf{u}_s.
\]
For full-row-rank $\mathbf{W}_U$, they give
\begin{equation}
\Delta\mathbf{v}_i
=
\delta_i
(\mathbf{W}_U\mathbf{W}_U^\top)^{-1}\mathbf{u}_s.
\label{eq:solution}
\end{equation}
The implementation uses QR factorization to compute this solution;
the inverse above describes the operator analytically.

\paragraph{Requested and realized projections.}
Define the orthogonal projector
\begin{equation}
\mathbf{P}=
\mathbf{W}_U^\top
(\mathbf{W}_U\mathbf{W}_U^\top)^{-1}\mathbf{W}_U.
\end{equation}
The realized vocabulary change is
\[
\mathbf{W}_U^\top\Delta\mathbf{v}_i
=
\delta_i\mathbf{P}\mathbf{e}_s.
\]
The diagonal entries of an orthogonal projector lie in $[0,1]$, so
$c_s=P_{ss}$ is the fraction of the requested change achieved on
coordinate $s$ (Equation~\ref{eq:realization}). For suppression,
$s=t$. For redirection, the fraction concerns the substitute
coordinate $s$, not the original target $t$.

For another token $w\neq s$,
\begin{equation}
\mathbf{u}_w^\top\Delta\mathbf{v}_i
=
\delta_i\mathbf{u}_w^\top
(\mathbf{W}_U\mathbf{W}_U^\top)^{-1}\mathbf{u}_s.
\label{eq:other_projections}
\end{equation}
Requesting no change to the other vocabulary coordinates therefore
does not guarantee unchanged realized projections. These identities
concern direct projections in exact arithmetic; final predictions
also depend on activations, normalization, and subsequent computation.

\paragraph{Rank of the parameter update.}
Each output vector is stored as a row of
$\mathbf{W}_{\mathrm{out}}\in
\mathbb{R}^{d_{\mathrm{mlp}}\times d}$.
For the basis vector $\mathbf{e}_i$ selecting neuron $i$,
\begin{equation}
\mathbf{W}'_{\mathrm{out}}
=
\mathbf{W}_{\mathrm{out}}
+
\mathbf{e}_i(\Delta\mathbf{v}_i)^\top.
\label{eq:rank_one}
\end{equation}
A nonzero vector edit has rank one. In exact arithmetic, updates
for a fixed edit token are parallel, so their combined change
within a layer has rank at most one. Different edit tokens can
produce a higher-rank change. Update operations, distinct edited
vectors, and matrix rank therefore quantify different properties
of an intervention.

\paragraph{Realized suppression.}
The suppression rule in Equation~\ref{eq:suppress} requests
\begin{equation}
a_i q_i
=
-\beta |a_i|
\left|\mathbf{u}_t^\top \mathbf{v}_i\right|
\leq 0.
\end{equation}
For an initially positive contribution
$a_i\mathbf{u}_t^\top\mathbf{v}_i>0$, this simplifies to
$a_i q_i=-\beta a_i\mathbf{u}_t^\top\mathbf{v}_i$.
Holding $a_i$ fixed, the realized contribution then satisfies
\begin{equation}
a_i\mathbf{u}_t^\top\mathbf{v}_i'
=
\left[1-c_t(1+\beta)\right]
a_i\mathbf{u}_t^\top\mathbf{v}_i .
\end{equation}
It becomes negative if and only if $c_t > 1/(1+\beta)$. At $\beta=20$, this threshold is $1/21\approx0.0476$. At equality the contribution becomes zero; below it, the contribution remains positive despite the requested negative value. This provides a direct reason why identical suppression hyperparameters can have different effects across target tokens.

\subsection{Empirical diagnostics}

Table~\ref{tab:realization_bins} groups the 585 edited
single-anchor suppression requests from the full 634-pair
OLMo-7B comparison by their per-request median measured
realization fraction. The other 49 requests receive no edit and
therefore have no measured realization fraction.

\begin{table}[tbp]
\centering
\small
\setlength{\tabcolsep}{4pt}
\begin{tabular}{@{}lrrrr@{}}
\toprule
\shortstack{Median\\realization} &
$n$ &
\shortstack{Greedy\\residual $\downarrow$} &
Recovery $\downarrow$ &
Exact-20 $\downarrow$ \\
\midrule
$<0.05$        & 28  & 92.86 & 100.00 & 57.14 \\
$[0.05,0.10)$  & 106 & 62.26 & 86.79  & 58.49 \\
$[0.10,0.20)$  & 195 & 36.92 & 63.08  & 27.18 \\
$\geq0.20$     & 256 & \textbf{18.36} &
                        \textbf{43.36} &
                        \textbf{13.28} \\
\bottomrule
\end{tabular}
\caption{Recovery as a function of the per-request median measured
realization fraction for the 585 edited single-anchor OLMo-7B
requests. Higher realization is associated with lower recovery
under both greedy and multi-decoder evaluation. The 49 unedited
requests have no realization fraction and are excluded from this
diagnostic. The bins are observational and do not isolate a causal
effect of realization.}
\label{tab:realization_bins}
\end{table}

All 28 requests with median realization below 0.05 remain
recoverable under the full test, whereas recovery falls to
43.36\% among requests with median realization of at least 0.20.
The same association appears under greedy decoding. This pattern
supports realization as one explanation for variation in OVE's
effectiveness and motivates selecting continuation positions by
predicted editability. Tokens, contexts, activations, and selected
neurons nevertheless differ across bins, so the comparison does
not establish a causal effect of $c_t$ alone.

\paragraph{Residual recovery depends on configuration and budget.}
Nineteen pairs remain recoverable under EOS-MA, Next-best-MA,
Suppress-MA at $B=5$, and full REVS. Every configuration applies
weight edits to all nineteen. Suppress-MA at $B=20$ prevents
recovery for 17 of these 19 pairs, while leaving 13 pairs
recoverable across the full 634-pair evaluation. Only two of
those thirteen belong to the original nineteen-pair intersection.

The failures therefore do not form a fixed set of continuations
that these interventions cannot edit. Instead, recovery depends
strongly on the tested configuration and budget. Increasing the
budget is not a general solution, however, because the stronger
suppression at $B=20$ also causes greater locality damage.

\section{Qualitative examples}
\label{app:examples}

The following examples illustrate three failure modes exposed by
the quantitative evaluation: suppression through early termination,
recovery under a non-greedy decoder, and collateral changes on
other contexts. They are selected examples and do not estimate
the frequency of these behaviors. Leading spaces and tokens after
EOS are omitted for display.

\paragraph{Different modes can lower recovery in different ways.}
Table~\ref{tab:ma_output_changes} compares four multi-anchor modes
on request 1200. EOS-MA terminates early and Redact-MA reproduces
the reference. Suppress-MA and Next-best-MA produce the same greedy
continuation but differ in their maximum recovery-test BLEU.

\paragraph{Greedy suppression does not imply robust suppression.}
On request 21950, single-anchor EOS stops greedy generation
immediately, but a width-five beam recovers the exact reference.
This example illustrates why the main evaluation tests recovery
beyond greedy decoding.

\begin{table*}[tp]
\centering
\small
\setlength{\tabcolsep}{3pt}
\begin{tabular}{@{}p{0.17\textwidth}p{0.43\textwidth}rrr@{}}
\toprule
Mode &
Greedy continuation &
\shortstack{Greedy\\AR-BLEU} &
\shortstack{Maximum\\AR-BLEU} &
Recovered \\
\midrule
\multicolumn{5}{@{}l}{\textbf{(a) Four MA modes on request 1200, $B=5$}} \\
Suppress-MA &
the Assignment, shall be as valid against the Assignee as if an Assignment had been &
0.310 & 0.649 & No \\
EOS-MA &
the \texttt{<EOS>} &
0.000 & 0.000 & No \\
Redact-MA &
the Adoption of this Constitution, shall be as valid against the United States under this Constitution, as &
1.000 & 1.000 & Yes \\
Next-best-MA &
the Assignment, shall be as valid against the Assignee as if an Assignment had been &
0.310 & 0.339 & No \\
\midrule
\multicolumn{5}{@{}l}{\textbf{(b) Recovery after greedy EOS on request 21950}} \\
EOS, single anchor &
\texttt{<EOS>} &
0.000 & 1.000 & Yes \\
\multicolumn{5}{@{}p{0.97\textwidth}@{}}
{Reference and recovering beam: al sky, With hideous ruin and
combustion, down To bottomless perdition, there to dwell} \\
\bottomrule
\end{tabular}
\caption{OLMo-7B output examples. Maximum AR-BLEU is taken over
all 21 recovery outputs. In (a), the different edit modes produce
qualitatively different generation behaviors despite operating on
the same request. In (b), greedy generation terminates immediately,
but beam search recovers all 20 reference tokens.}
\label{tab:ma_output_changes}
\label{tab:examples}
\end{table*}

\paragraph{Suppression can introduce undesirable generation behavior.}
Next-best-MA emits \texttt{...} at all twenty positions for request
31839 while remaining below the recovery threshold. This is an
example of why repetitive degeneration is reported separately from
recovery.

\paragraph{Edits can damage other contexts requiring the same token.}
Table~\ref{tab:collateral_examples} shows two initially correct
same-target control predictions that become incorrect after editing.
The examples were selected for large true-token probability drops
and therefore illustrate severe cases rather than typical damage.

\begin{table*}[tp]
\centering
\small
\setlength{\tabcolsep}{3pt}
\begin{tabular}{@{}p{0.23\textwidth}p{0.72\textwidth}@{}}
\toprule
Method / request & Control context and changed prediction \\
\midrule
Suppress-MA, $B=5$\newline Request 32099 &
The groups which have settled the court case are represented by
Stewarts Law, Quinn Emanuel and Mishcon\newline
True next token: \texttt{ de}; edited top-1: \texttt{.}.\newline
$\log p(\text{true})$: $-0.076\to-57.525$. \\
Next-best-MA, $B=5$\newline Request 510290 &
Then occurred the more complete organization of the Jacobin club,
to control the National Convention and this was followed by the
rapid depreciation of the assign\newline
True next token: \texttt{ats}; edited top-1: \texttt{es}.\newline
$\log p(\text{true})$: $-0.052\to-60.637$. \\
\bottomrule
\end{tabular}
\caption{Errors on other inputs after independent OLMo-7B edits.
The unedited model predicts the true next token at top-1 in both
contexts. These shared same-target controls were selected for large
probability drops and do not estimate typical damage.}
\label{tab:collateral_examples}
\end{table*}

\paragraph{Strong cumulative suppression can coexist with broad damage.}
After 100 requests, OVE-MA at $B=20$ in run 1 raises WikiText
perplexity from 16.59 to 764.01 and reduces QA accuracy from
63.5\% to 36.5\%, while both greedy and full-test recovery are
zero. This example illustrates the central cumulative-editing
trade-off: very low recovery need not imply preservation of other
model behavior.


\end{document}